\documentclass{article} 

\usepackage{felix}

\begin{document}
\title{Replicable Clustering\thanks{Authors are listed alphabetically.}}
\author{%
  Hossein Esfandiari \\
  Google Research \\
  \texttt{esfandiari@google.com} \\
  \And
  Amin Karbasi \\
  Yale University, Google Research \\
  \texttt{amin.karbasi@yale.edu} \\
  \And
  Vahab Mirrokni \\
  Google Research \\
  \texttt{mirrokni@google.com} \\
  \And
  Grigoris Velegkas \\
  Yale University \\
  \texttt{grigoris.velegkas@yale.edu} \\
  \And
  Felix Zhou \\
  Yale University \\
  \texttt{felix.zhou@yale.edu}
}
\maketitle
\begin{abstract}
    We design replicable algorithms
    in the context of statistical clustering under the recently introduced notion
    of replicability from \citet{impagliazzo2022reproducibility}.
    According to this definition, a clustering algorithm is replicable if,
    with high probability,
    its output induces the \emph{exact} same partition of the sample space
    after two executions on different inputs drawn from the same distribution,
    when its \emph{internal} randomness
    is shared across the executions.
    We propose such algorithms for the statistical $k$-medians,
    statistical $k$-means,
    and statistical $k$-centers problems by utilizing approximation routines
    for their combinatorial counterparts in a black-box manner.
    In particular,
    we demonstrate a replicable $O(1)$-approximation algorithm for statistical Euclidean $k$-medians ($k$-means)
    with $\tilde O(\poly(k, d)k^{\log\log k})$ sample complexity.
    We also describe an $O(1)$-approximation algorithm
    with an additional $O(1)$-additive error for statistical Euclidean $k$-centers, albeit
    with $\tilde O(\poly(k)\exp(d))$ sample complexity.
    In addition,
    we provide experiments on synthetic distributions in 2D using the $k$-means++ implementation from sklearn as a black-box that 
    validate our theoretical results\footnote{\url{https://anonymous.4open.science/r/replicable_clustering_experiments-E380}}.
\end{abstract}

\section{Introduction}\label{sec:introduction}
The unprecedented increase in the amount of data that is available to
researchers across many different scientific areas has led to
the study and development of automated data analysis methods.
One fundamental category of such methods is \emph{unsupervised learning}
which aims to identify some inherent structure in \emph{unlabeled} data.
Perhaps the most well-studied way to do that is by grouping together data that share similar characteristics.
As a result, \emph{clustering} algorithms have become one of the central
objects of study in unsupervised learning.
Despite a very long line of work studying such algorithms,
e.g. \citet{jain1988algorithms, hart2000pattern,anderberg2014cluster},
there is not an agreed-upon definition that quantifies the quality of a clustering solution. \citet{kleinberg2002impossibility} showed that there is an inherent reason why this is the case:
it is impossible to design a clustering function that satisfies three natural properties, namely \emph{scale-invariance}, \emph{richness of solutions}, and  \emph{consistency}.
This means that the algorithm designer needs to balance several conflicting desiderata.
As a result, the radically different approaches that scientists use depending
on their application domain can be sensitive to several factors such as their random
initialization, the measure of similarity of the data, the presence of noise in the measurements,
and the existence of outliers in the dataset.
All these issues give rise to algorithms whose results are not
\emph{replicable}, i.e., when we execute them on two different
samples of the same population,
they output solutions that vary significantly.
This begs the following question. Since the goal of clustering
algorithms is to reveal properties of the underlying population,
how can we trust and utilize their results when they fail to pass this simple test?

Replicability is imperative in making sure that scientific findings are both valid and reliable.
Researchers have an obligation to provide coherent results and conclusions across multiple repetitions of the same experiment.
Shockingly,
a 2016 survey that appeared in Nature \citep{baker20161} revealed that 70\% of researchers tried, but were unable to, replicate the findings of another researcher
and more than 50\% of them believe there is a significant crisis in replicability.
Unsurprisingly,
similar worries have been echoed in the subfields of machine learning and data science \citep{pineau2019iclr,pineau2021improving}.
In this work,
we initiate the study of replicability in clustering,
which is one of the canonical problems of unsupervised learning.

\subsection{Related Works}
\textbf{Statistical Clustering.}
The most relevant previous results for (non-replicable) statistical clustering was established by \citet{ben2007framework},
who designed $O(1)$-approximation algorithms for statistical $k$-medians ($k$-means) with $O(k)$ sample complexity.
However,
their algorithm picks centers from within the samples and is therefore non-replicable.

\textbf{Combinatorial Clustering.}
The flavor of clustering most studied in the approximation algorithms literature
is the setting where we have a uniform distribution over finite points
and the algorithm has explicit access to the entire distribution.
Our algorithms rely on having black-box access to a combinatorial clustering oracle.
See \citet{byrka2017improved, ahmadian2019better} for the current best polynomial-time approximation algorithms for combinatorial $k$-medians ($k$-means) in general metrics with approximation ratio 2.675 (9).
Also, see \citet{cohen2022improved} for a $2.406$ ($5.912$) approximation algorithm for the combinatorial Euclidean $k$-medians ($k$-means).

\textbf{Clustering Stability.} Stability in clustering has been studied both from a practical
and a theoretical point of view \citep{ben2001stability, lange2004stability, von2005towards, ben2006sober, rakhlin2006stability, ben2007stability, von2010clustering}.
In most applications,
it is up to the 
algorithm designer to decide upon the value of $k$,
i.e., the number of different clusters.
Thus, it was proposed that a necessary condition it
should satisfy is that it leads to solutions that are not 
very far apart under \emph{resampling} of the input data
\citep{ben2001stability, lange2004stability}.
However,
it was shown that this notion of stability for center-based clustering
is heavily based on symmetries within the data
which may be unrelated to clustering parameters \citep{ben2006sober}.
Our results differ from this line of work in that we require the output
across two separate samples to be \emph{exactly the same} with high probability,
when the randomness is shared.
Moreover, our work reaffirms \citet{ben2006sober} in that their notion of stability
can be perfectly attained no matter the choice of $k$.

Other notions of stability related to our work include
robust hierarchical clustering \citep{balcan2014robust},
robust online clustering \citep{lattanzi2021robust},
average sensitivity \citep{yoshida2022average},
and differentially private (DP) clustering \citep{cohen2021differentially, ghazi2020differentially}.
The definition of replicability we use is statistical and relies on an underlying data distribution
while (DP) provides a worst-case combinatorial guarantee for two runs of the algorithm
on neighboring datasets.
\citet{bun2023stability, kalavasis2023statistical} provide connections between DP and replicability for statistical learning problems. However, these transformations are not computationally
efficient. It would be interesting to come up with
computationally efficient reductions between replicable and DP
clustering algorithms.

\textbf{Coresets for Clustering.} A long line of work has focused on developing strong coresets
for various flavors of centroid-based clustering problems.
See \citet{sohler2018strong} for an overview of this rich line of work.
The most relevant for our results
include coresets for dynamic geometric streams through hierarchical grids \citep{frahling2005coresets}
and sampling based techniques \citep{ben2007framework, feldman2011unified, bachem2018scalable}.

\textbf{Dimensionality Reduction.} Dimensionality reduction for clustering has been a popular area of study
as it reduces both the time and space complexity of existing algorithms.
The line of work on data-oblivious dimensionality reduction for $k$-means clustering was initiated by \citet{boutsidis2010random}.
The goal is to approximately preserve the cost of all clustering solutions
after passing the data through a dimensionality reduction map.
This result was later improved and generalized to $(k, p)$-clustering \citep{cohen2015dimensionality,becchetti2019oblivious},
culminating in the work of \citet{makarychev2019performance},
whose bound on the target dimension is sharp up to a factor of $\log \nicefrac1\varepsilon$.
While \citet{charikar2022johnson} overcome this factor,
their result only preserves the cost across the optimal solution.

\textbf{Replicability in ML.} Our results extend the recently initiated
line of work on designing provably replicable learning algorithms under
the definition that was introduced by
\citet{impagliazzo2022reproducibility}.
Later, \citet{esfandiari2022reproducible} considered a natural adaption
of this definition to the setting of bandits and designed 
replicable algorithms that have small regret.
A slightly different notion of replicability in optimization was 
studied in \citet{ahn2022reproducibility}, where it is required that
an optimization algorithm that uses noisy operations during its execution,
e.g., noisy gradient evaluations,
outputs solutions that are close when executed twice.
Subsequently, \citet{bun2023stability, kalavasis2023statistical} established
strong connections between replicability and other notions
of algorithmic stability. Recently, \citet{dixon2023list, chase2023replicability} proposed a weaker notion of 
replicability where the algorithm is not required to output
the same solution across two executions, but its output needs
to fall into a small list of 
solutions.

\section{Setting \& Notation}
Let $\mcal X\sset \R^d$ be the instance space endowed with a 
metric $\kappa: \mcal X\times \mcal X\to \R_+$
and $\P$ be a distribution on $\mcal X$ which
generates the i.i.d. samples that the learner observes.

For $F \subseteq \R^d$ and $x\in \mcal X$,
we overload the notation and write $F(x) := \argmin_{f\in F} \kappa(x, f)$
to be the closest point to $x$ in $F$
as well as $\kappa(x,F) := \kappa(x,F(x))$
to be the shortest distance from $x$ to a point in $F$.

We assume that $\mcal X$ is a subset of $\mcal B_d$,
the $d$-dimensional $\kappa$-ball of diameter 1 centered about the origin\footnote{Our results can be generalized to $\kappa$-balls of diameter $L$ with an arbitrary center through translation and scaling.}.
We also assume that $\kappa$ is induced by some norm $\norm{\cdot}$ on $\R^d$
that is \emph{sign-invariant} (invariant to changing the sign of a coordinate)
and \emph{normalized} (the canonical basis has unit length).
Under these assumptions,
the unit ball of $\kappa$ is a subset of $[-1, 1]^d$.

Our setting captures a large family of norms,
including the $\ell_p$-norms,
Top-$\ell$ norms (sum of $\ell$ largest coordinates in absolute value),
and ordered norms (non-negative linear combinations of Top-$\ell$ norms) \citep{chakrabarty2019approximation}.
Our results hold for more general classes of norms
but for the sake of simplicity,
we abide by these assumptions.

We define $\Delta := \sup\set{\kappa(x, y): x, y\in [0, 1)^d}$
to be the $\kappa$-diameter of the unit hypercube.
Note that $1\leq \Delta \leq d$ by assumption.
Moreover,
$L \Delta$ is the $\kappa$-diameter of a hypercube with side length $L$.
For example,
if $\norm{\cdot} = \norm{\cdot}_2$ is the Euclidean norm,
then $\Delta = \sqrt d$.

\subsection{Clustering Methods and Generalizations}
We now introduce the clustering objectives
that we study in this work,
which all fall in the category of minimizing a cost function $\cost: \mcal F\to \R_+$,
where $\mcal F := \set{F\sset \mcal B_d: \card F = k}$.
We write $\cost(F)$ to denote the objective in the statistical setting
and $\widehat\cost(F)$ for the combinatorial setting
in order to distinguish the two.

\begin{prob}[Statistical $(k, p)$-Clustering]\label{prob:stat cluster}
  Given i.i.d. samples from a distribution $\P$ on $\mcal X\sset \mcal B_d$,
  minimize $\cost(F) := \E_{x\sim \P} \kappa(x, F)^p$.
\end{prob}
In other words,
we need to partition the points into $k$
clusters so that the expected distance of a point to 
the center of its cluster, measured by $\kappa(\cdot,\cdot)^p$,
is minimized.
This is closely related to the well-studied
combinatorial variant of the $(k,p)$-clustering problem.

\begin{prob}[$(k, p)$-Clustering]\label{prob:cluster}
  Given some points $x_1, \dots, x_n\in \mcal X$,
  minimize $\widehat \cost(F) := \frac1n \sum_{i=1}^n \kappa(x_i, F)^p$.
\end{prob}

We note that (statistical) $k$-medians and (statistical) $k$-means is a special case of \Cref{prob:cluster} (\Cref{prob:stat cluster})
 with $p=1, 2,$ respectively. We also consider a slight
variant of the combinatorial problem, i.e., \Cref{prob:cluster},
where we allow different points $x_i$ to participate with different weights $w_i$
in the objective. We refer to this problem as the \emph{weighted} $(k, p)$-clustering problem.

We now shift our attention to the $k$-centers problem. 

\begin{prob}[Statistical $k$-Centers]\label{prob:stat k-centers}
  Given i.i.d. samples from a distribution $\P$ on $\mcal X\sset \mcal B_d$,
  minimize $\cost(F) := \max_{x \in \mcal X} \kappa(x, F)$.
\end{prob}

Notice that the $\ell_\infty$ norm is the limit of the $\ell_p$ norm as $p$ tends to infinity,
hence $k$-centers is,
in some sense,
the limit of $(k, p)$-clustering as $p$ tends to infinity.
Also,
notice that this problem differs from $k$-means and $k$-medians in the sense that it has a \emph{min-max} flavor,
whereas the other two are concerned with minimizing some \emph{expected} values.
Due to this difference, we need to treat $k$-centers separately from the other two problems, and we need to make some assumptions in order
to be able to solve it from samples (cf. \Cref{assum:clusterable}, \Cref{assum:reproducible}). We elaborate more on that later.

Let us also recall the combinatorial version of $k$-centers.

\begin{prob}[$k$-Centers]\label{prob:k-centers}
  Given some points $x_1, \dots, x_n\in \mcal X$,
  minimize $\widehat\cost(F) := \max_{i \in [n]} \kappa(x_i, F)$.
\end{prob}


We remark that clustering has mainly been studied from the combinatorial point of view,
where the distribution is the uniform distribution over some finite points 
and we are provided the entire distribution.
The statistical clustering setting generalizes to arbitrary distributions with only sample access.
We emphasize that although we only have access to samples,
our output should be a good solution for the entire distribution
and not just the observed data.

We write $F_{\OPT}$ to denote an optimal solution for the entire distribution
and $\OPT := \cost(F_{\OPT})$. 
Similarly,
we write $\wh F_{\OPT}$ to denote an optimal sample solution
and $\wh\OPT := \wh\cost(\wh F_{\OPT})$.
Suppose we solve \Cref{prob:k-centers} given a sample of size $n$ from \Cref{prob:stat k-centers}.
Then $\wh \OPT \leq \OPT$ since we are optimizing over a subset of the points.

Recall that a \emph{$\beta$-approximate solution} $F$ is one which has cost $\cost(F) \leq \beta \OPT$.
Note this is with respect to the statistical version of our problems.
An algorithm that outputs $\beta$-approximate solutions is known as a \emph{$\beta$-approximation algorithm}.
We also say that $F$ is a \emph{$(\beta, B)$-approximate solution} if $\cost(F)\leq \beta \OPT + B$.


\subsubsection{\texorpdfstring{Parameters $p$ and $\kappa$}{Parameters p and kappa}}
Here we clarify the difference between $p$ and $\kappa$,
which are two separate entities in the cost function $\E_x \left[ \kappa(x, F(x))^p \right]$.
We denote by $\kappa$ the distance metric used to measure the similarity between points.
The most commonly studied and applied option is the Euclidean distance
for which our algorithms are the most sample-efficient.
On the other hand,
$p$ is the exponent to which we raise the distances
when computing the cost of a clustering.
A smaller choice of $p$ puts less emphasis on points that are far away from centers
and $p=1$ seeks to control the average distance to the nearest center.
A large choice of $p$ puts emphasis on points that are further away from centers
and as $p$ tends to infinity,
the objective is biased towards solutions minimizing the maximum distance to the nearest center.
Thus we can think of $k$-centers as $(k, p)$-clustering when $p=\infty$.
As a concrete example,
when $\kappa$ is the Euclidean distance
and $p=5$,
the cost function becomes $\E_x\left[ \norm{x-F(x)}_2^5 \right]$.

\subsection{Replicability}
Throughout this work, we study replicability\footnote{Originally this definition was 
called reproducibility \citep{impagliazzo2022reproducibility} but it was later pointed out that the correct term is replicability \citep{ahn2022reproducibility}.} as an algorithmic property using the definition of \citet{impagliazzo2022reproducibility}.

\begin{defn}[Replicable Algorithm; \citep{impagliazzo2022reproducibility}]
  Let $\rho \in (0,1)$. A randomized algorithm $\mcal A$ is \emph{$\rho$-replicable} if
  for two sequences of $n$ i.i.d. samples $\bar X$,
  $\bar Y$ generated from some distribution $\P^n$
  and a random binary string $\bar r \sim R(\mcal X)$,
  \[
    \P_{\bar X, \bar Y \sim \P^n, \bar r \sim R(\mcal X)} \set{\mcal A(\bar X; \bar r) = \mcal A(\bar Y; \bar r)}
    \geq 1-\rho ,.
  \]
\end{defn}

In the above definition, we treat $\mcal A$ as a randomized mapping
to solutions of the clustering problem.
Thus, even when $\bar X$ is fixed, $\mcal A(\bar X)$ should be thought of as random variable,
whereas $\mcal A(\bar X; \bar r)$ is the \emph{realization}
of this variable given the (fixed) $\bar X, \bar r.$
We should think of $\bar r$ as the shared randomness between the two executions.
In practice, it can be implemented as a shared random seed. 
We underline that sharing the randomness across executions is crucial 
for the development of our algorithms.
We also note that by doing that we \emph{couple} the two random variables $\mcal A(\bar X), \mcal A(\bar Y)$, whose realization depends on $r \sim R(\mcal X)$.
Thus, if their realizations
are equal with high probability under this coupling,
it means that the distributions of $\mcal A(\bar X), \mcal A(\bar Y)$ are
\emph{statistically close}. This connection is discussed further
 in \citet{kalavasis2023statistical}.

In the context of a clustering algorithm $\mcal A$,
we interpret the output $\mcal A(\bar X; \bar r)$
as a clustering function $f:\mcal X\to [k]$ which partitions the support of $\P$. The definition of
$\rho$-replicability demands that $f$ is the same with probability at least $1-\rho$ across two executions.
We note that in the case of centroid-based clustering
such as $k$-medians and $k$-means,
the induced partition is a function of the centers
and thus it is sufficient to output the exact same centers with probability $1-\rho$ across two executions.
However, we also allow for algorithms that create partitions implicitly
without computing
their centers explicitly.

Our goal is to develop replicable clustering algorithms for $k$-medians, $k$-means, and $k$-centers,
which necessitates that the centers we choose are arbitrary points within $\R^d$
and \emph{not} only points among the samples. We underline that as
in the case of differential privacy, it is trivial
to design algorithms that satisfy
the replicability property,
e.g. we can let $\mcal A$ be the constant mapping.
The catch is that these algorithms do not achieve any \emph{utility}.
In this work, we are interested in
designing replicable clustering algorithms whose utility is competitive with
their non-replicable counterparts.

\section{Main Results}
In this section,
we informally state our results for replicable statistical $k$-medians ($(k, 1)$-clustering),
$k$-means ($(k, 2)$-clustering),
and $k$-centers under general distances.
Unfortunately,
generality comes at the cost of exponential dependency on the dimension $d$.
We also state our results for replicable statistical $k$-medians
and $k$-means specifically under the Euclidean distance,
which has a polynomial dependency on $d$.
Two key ingredients is the uniform convergence of $(k, p)$-clustering costs (cf. \Cref{thm:uniform convergence})
as well as a data-oblivious dimensionality reduction technique for $(k, p)$-clustering
in the distributional setting (cf. \Cref{thm:dim reduction}).
These results may be of independent interest.

We emphasize that the Euclidean $k$-median and $k$-means
are the most studied and applied flavors of clustering,
thus the sample complexity for the general case
and the restriction to $p=1, 2$
does not diminish the applicability of our approach.

The main bottleneck in reducing the sample complexity for general norms
is the lack of a data-oblivious dimensionality reduction scheme.
This bottleneck is not unique to replicability
and such a scheme for general norms would be immediately useful for many distance-based problems
including clustering.
It may be possible to extend our results to general $(k, p)$-clustering beyond $p=1, 2$.
The main challenge is to develop an approximate triangle inequality for $p$-th powers of norms.
Again,
this limitation is not due to replicability
but rather the technique of hierarchical grids.
It is a limitation shared by \citet{frahling2005coresets}.

Before stating our results,
we reiterate that the support of our domain $\mcal X$
is a subset of the unit-diameter $\kappa$-ball $\mcal B_d$.
In particular,
we have that $\OPT \leq 1$.

\begin{thm}[Informal]\label{thm:statistical clustering informal}
  Let $\varepsilon, \rho\in (0, 1)$.
  Given black-box access to a $\beta$-approximation oracle for weighted $k$-medians,
  respectively weighted $k$-means
  (cf. \Cref{prob:cluster}),
  there is a $\rho$-replicable algorithm for statistical $k$-medians,
  respectively $k$-means (cf. \Cref{prob:stat cluster}),
  such that with probability at least $0.99$,
  it outputs a $(1+\varepsilon)\beta$-approximation.
  Moreover,
  the algorithm has sample complexity
  \[
    \tilde O \left( \poly\left( \frac{k}{\rho \OPT} \right) \left( \frac{2\Delta}{\varepsilon} \right)^{O(d)} \right) \,.
  \]
\end{thm}
When we are working in Euclidean space, we can get improved results
for these problems.
\begin{thm}[Informal]\label{thm:euclidean clustering informal}
  Let $\rho\in (0, 1)$.
  Suppose we are provided with black-box access 
  to a $\beta$-approximation oracle for weighted \emph{Euclidean} $k$-medians ($k$-means).
  Then there is a $\rho$-replicable algorithm that partitions the input space
  so with probability at least $0.99$,
  the cost of the partition is at most $O(\beta \OPT)$.
  Moreover,
  the algorithm has sample complexity
  \[
    \tilde O \left( \poly\left( \frac{d}{\OPT} \right) \left( \frac{k}\rho \right)^{O(\log\log (\nicefrac{k}\rho))} \right) \,.
  \]
\end{thm}
We underline that in this setting we compute 
an \emph{implicit} solution to \Cref{prob:stat cluster},
since we do not output $k$ centers.
Instead, we output a function $f$ that takes as input
a point $x \in \mcal X$ and outputs the label of the cluster it
belongs to in polynomial time. The replicability guarantee states
that, with probability $1-\rho$, the function will be the
same across two executions.

The combinatorial $k$-medians ($k$-means) problem where the centers are
restricted to be points of the input is a well-studied problem
from the perspective of polynomial-time constant-factor approximation algorithms.
See \citet{byrka2017improved, ahmadian2019better} for the current best polynomial-time approximation algorithms for combinatorial $k$-medians ($k$-means) in general metrics with approximation ratio 2.675 (9).
Also, see \citet{cohen2022improved} for a $2.406$ ($5.912$) approximation algorithm for the combinatorial Euclidean $k$-medians ($k$-means).

As we alluded to before, in order to solve $k$-centers from samples,
we need to make an additional assumption. Essentially,
\Cref{assum:reproducible} states that there is
a $(\beta, B)$-approximate solution $F$, such that, with some constant probability, e.g. $0.99$,
when we draw $n$ samples from $\P$
we will observe at least one sample from each cluster of $F$.
\begin{thm}[Informal]\label{thm:informal k-centers sample complexity of reproducible active cells}
  Let $c\in (0, 1)$.
  Given black-box access to a $(\hat{\beta}, \hat{B})$-approximation oracle for $k$-centers (cf. \Cref{prob:k-centers}) and under \Cref{assum:reproducible}, there is 
  a $\rho$-replicable algorithm for statistical $k$-centers (cf. \Cref{prob:stat k-centers}),
  that outputs a $(O(\beta + \hat{\beta}), O(B + \wh B +  (\beta + \hat \beta + 1) c) \Delta)$-approximate solution
  with probability at least $0.99$.
  Moreover, it
  has sample complexity
  \[
      \tilde O\left(\frac{n^2k \left(\nicefrac{1}{c}\right)^{3d}}{\rho^2q^2} \right) \,.
  \]
\end{thm}
Recall that there is a simple greedy 2-approximation for the sample $k$-center problem
whose approximation ratio cannot be improved unless P = NP \citep{hochbaum1985best}.

We defer the discussion around $k$-centers to \Cref{sec:overview k-centers}.
In particular,
see \Cref{thm:formal k-centers sample complexity of reproducible active cells} in \Cref{sec:overview k-centers}
for the formal statement of \Cref{thm:informal k-centers sample complexity of reproducible active cells}.

\section{Overview of \texorpdfstring{$(k, p)$}{(k, p)}-Clustering}
In this section, we present our approach to the $(k,p)$-clustering problem.
First,
we replicably approximate the distribution with a finite set of points
by extending the approach of \citet{frahling2005coresets}
to the distributional setting.
Then, we solve the combinatorial $(k, p)$-clustering problem on this coreset
using an approximation oracle in a black-box manner.
In the following subsections, we give a more detailed
overview for each step of our approach.
For the full proofs and technical details,
we kindly refer the reader to \Cref{sec:multinomial estimation} - \ref{sec:statistical clustering summary}.
In summary:
\begin{enumerate}[1)]
  \item Replicably build a variant of a \emph{quad tree} \citep{finkel1974quad} (cf. \Cref{sec:replicable quad tree}).
  \item Replicably produce a weighted \emph{coreset} using the quad tree (cf. \Cref{sec:coresets}).
  \item Apply the optimization oracle for the combinatorial problem on the coreset.
\end{enumerate}

For general norms,
this approach leads to an exponential dependence on $d$.
However,
we are able to handle the case of Euclidean distances
by extending existing dimensionality reduction techniques for sample Euclidean $(k, p)$-clustering \citep{makarychev2019performance}
to the distributional case (cf. \Cref{sec:euclidean metric}).
Thus, for the widely used Euclidean norm,
our algorithm has $\poly(d)$ sample complexity.

\subsection{Coresets}\label{sec:coresets}
\begin{defn}[(Strong) Coresets]
    For a distribution $\P$ with support $\mcal X\sset \mcal B_d$
    and $\varepsilon\in (0, 1)$,
    a \emph{(strong) $\varepsilon$-coreset} for $\mcal X$
    is a distribution $\P'$ on $\mcal X'\sset B_d$ which satisfies
    \begin{align*}
      (1-\varepsilon) \cdot \E_{x\sim \P} \kappa(x, F)^p  
      &\leq \E_{x'\sim \P'} \left[ \kappa(x', F)^p \right]
      \leq (1+\varepsilon) \cdot \E_{x\sim \P} \kappa(x, F)^p
    \end{align*}
    for every set of centers $F\sset \mcal B_d, \card F=k$.
\end{defn}
Essentially,
coresets help us approximate the true cost 
on the distribution $\P$ by considering another distribution $\P'$ 
whose support $\mcal X'$ can be arbitrarily smaller than the support of $\P$.

Inspired by \citet{frahling2005coresets},
the idea is to replicably consolidate our distribution $\P$ through some mapping $R: \mcal X\to \mcal X$
whose image has small cardinality $\card{R(\mcal X)} << \infty$
so that for any set of centers $F$,
\begin{align*}
  (1-\varepsilon) \E_x \kappa(x, F)^p
  &\leq \E_x \kappa(R(x), F)^p
  \leq (1+\varepsilon) \E_x \kappa(x, F)^p \,,
\end{align*}
where $\varepsilon\in (0, 1)$ is some error parameter.
In other words,
$(\P_R, R(\mcal X))$ is an $\varepsilon$-coreset.
Note that given the function $R$,
we can replicably estimate the probability mass at each point in $R(\mcal X)$
and then apply a weighted $(k, p)$-clustering algorithm. 

\subsection{Replicable Quad Tree}\label{sec:replicable quad tree}
We now explain how to replicably obtain the mapping $R: \mcal X\to \mcal X$
by building upon the work of \citet{frahling2005coresets}.
The pseudocode of the approach is provided in \Cref{alg:replicable quad tree}.
While \citet{frahling2005coresets} present their algorithm using hierarchical grids,
we take an alternative presentation using the quad tree \citep{finkel1974quad}, which could be of independent interest.

First,
we recall the construction of a standard quad tree in dimension $d$.
Suppose we have a set of $n$ points in $[-\nicefrac12, \nicefrac12]^d$.
The quad tree is a tree whose nodes represent hypercubes containing points
and can be built recursively as follows:
The root represents the cell $[-\nicefrac12, \nicefrac12]^d$ and contains all points.
If a node contains more than one point,
we split the cell it represents into $2^d$ disjoint,
equally sized cells.
For each non-empty cell,
we add it as a child of the current node and recurse into it.
The recursion stops when a node contains only 1 point. 
In the distributional setting,
the stopping criterion is
either when the diameter of a node is less than some length
or when the node contains less than some mass,
where both quantities are a function of the depth of the node.
See \Cref{alg:replicable quad tree}.

A quad tree implicitly defines a function $R: \mcal X\to \mcal X$ as follows.
Given a point $x\in \mcal X$ and the root node of our tree,
while the current node has a child,
go to the child containing $x$ if such a child exists,
otherwise,
go to any child.
At a leaf node,
output the center of the cell the leaf represents.
Intuitively,
the quad tree consolidates regions of the sample space into single points.
The construction can be made replicable since the decision to continue the recursion or not
is the only statistical operation 
and is essentially a heavy-hitters operations which can be performed in a replicable fasion.

Let $\mcal G_i$ denote the union of all $2^{id}$ possible cells at the $i$-th level.
We write $\P_i$ to denote the discretized distribution to $\mcal G_i$.
In other words,
$\P_i = \P|_{\sigma(\mcal G_i)}$ is the restriction of $\P$
to the smallest $\sigma$-algebra containing $\mcal G_i$.
Moreover,
we write $\Lambda$ to denote a replicable estimate of $\OPT$
with relative error $\varepsilon$,
say $\nicefrac1{\beta (1+\varepsilon)} \OPT \leq \Lambda \leq (1+\varepsilon) \OPT$
for some absolute constant $\beta \geq 1$.
We demonstrate how to obtain such a replicable estimate in \Cref{sec:k-medians OPT estimation}.

\begin{algorithm}[h]
\caption{Replicable Quad Tree}\label{alg:replicable quad tree}
\begin{algorithmic}[1]
  \STATE {\bfseries rQuadTree}(distribution $\P$, accuracy $\varepsilon$, exponent $p$, replicability $\rho$, confidence $\delta$):
    \STATE Init the node on the first level $\mcal Z[0] \gets \set*{[-\nicefrac12, \nicefrac12]^d}$.
    \FOR {depth $i\gets 1$; $\mcal Z[i-1]\neq \varnothing$ AND $(2^{-i+1} \Delta)^p > \nicefrac{\varepsilon \Lambda}{5}$; $i\gets i+1$}
      \STATE \COMMENT{$\P_i$ is the discretized distribution over $2^{id}$ cells on the $i$-th layer}
      \STATE \COMMENT{$\gamma$ is a parameter to be determined later.}
      \STATE \COMMENT{$t$ is an upper bound on the number of layers.}
      \STATE Compute $\mcal H = \rHeavyHitters\left( \P_i, v=\frac{\gamma\cdot \Lambda}{2^{-pi}}, \frac{v}2, \frac\rho{t}, \frac\delta{t} \right)$
      \FOR {node $Z\in \mcal Z[i-1]$}
        \FOR {heavy hitter cells $H\in \mcal H$ such that $H\sset Z$}
          \STATE $\children(Z) \gets \children(Z)\cup \set{H}$
          \STATE $\mcal Z[i] \gets \mcal Z[i]\cup \set{H}$.
        \ENDFOR
      \ENDFOR
    \ENDFOR
    \STATE Output root node.
\end{algorithmic}
\end{algorithm}

Our technique differs from that of \citet{frahling2005coresets} in at least three ways.
Firstly,
they performed their analysis for finite, uniform distributions with access to the entire distribution,
while our results hold assuming only sample access to a general bounded distribution\footnote{Our results also generalize to unbounded distributions with sufficiently small tails}.
Secondly,
\citet{frahling2005coresets} bound the number of layers as a function of the cardinality of the support.
For us,
this necessitates the extra termination condition when the side lengths of our grids fall below a fraction of $\Lambda$ as our distribution may have infinite support.
Finally,
\citet{frahling2005coresets} estimate $\OPT$ by enumerating powers of 2.
This suffices for their setting since their distributions are discrete and bounded.
However,
we require a more nuanced approach (cf. \Cref{sec:k-medians OPT estimation}) as we do not have a lower bound for $\OPT$.
We tackle this by showing uniform convergence of the clustering solution costs, which we establish via metric entropy and Rademacher complexity (cf. \Cref{sec:uniform convergence}). 

\subsection{Putting it Together}
Once we have produced the function $R$ implicitly through a quad tree,
there is still the matter of extracting a replicable solution from a finite distribution.
We can accomplish this by replicably estimating the probability mass at each point of $R(\mcal X)$
and solving an instance of the weighted sample $k$-medians ($k$-means).
This leads to the following results.
For details,
see \Cref{sec:multinomial estimation} - \ref{sec:statistical clustering summary}.

\begin{restatable}[\Cref{thm:statistical clustering informal}; Formal]{thm}{kMedians}\label{thm:statistical k-medians algorithm}
  Let $\varepsilon, \rho\in (0, 1)$ and $\delta\in (0, \nicefrac\rho3)$.
  Given black-box access to a $\beta$-approximation oracle for weighted $k$-medians
  (cf. \Cref{prob:cluster}),
  there is a $\rho$-replicable algorithm for statistical $k$-medians (cf. \Cref{prob:stat cluster})
  such that, with probability at least $1-\delta$,
  it outputs a $(1+\varepsilon)\beta$-approximation.
  Moreover,
  it has sample complexity
  \[
    \tilde O\left( \left( \frac{k^2d^2}{\varepsilon^{12}\rho^6\cdot \OPT^{12}}
    + \frac{k^3 2^{18d} \Delta^{3d+3}}{\rho^2 \varepsilon^{3d+5}\cdot \OPT^3} \right)\log \frac1\delta \right) \,.
  \]
\end{restatable}

\begin{restatable}[\Cref{thm:statistical clustering informal}; Formal]{thm}{kMeans}\label{thm:statistical k-means algorithm}
  Given black-box access to a $\beta$-approximation oracle for weighted $k$-means
  (cf. \Cref{prob:cluster}),
  there is a $\rho$-replicable algorithm for statistical $k$-means (cf. \Cref{prob:stat cluster})
  such that, with probability at least $1-\delta$,
  it replicably outputs a $(1+\varepsilon)\beta$-approximation.
  Moreover,
  it has sample complexity
  \[
    \tilde O\left( \left( \frac{k^2d^2}{\varepsilon^{12}\rho^6\cdot \OPT^{12}}
    + \frac{k^3 2^{39d} \Delta^{3d+6}}{\rho^2 \varepsilon^{6d+8}\cdot \OPT^3}\right) \log\frac1\delta \right) \,.
  \]
\end{restatable}

\section{The Euclidean Metric, Dimensionality Reduction, and \texorpdfstring{$(k, p)$}{(k, p)}-Clustering}\label{sec:euclidean metric}
In this section,
we focus on the Euclidean metric $\kappa(y, z) := \norm{y-z}_2$
and show how the sample complexity for Euclidean $(k, p)$-clustering can be made polynomial in the ambient dimension $d.$
Note that results for dimensionality reduction exist for the combinatorial Euclidean $(k, p)$-clustering problem \citep{charikar2022johnson, makarychev2019performance}.
However,
these do not extend trivially to the distributional case.
We remark that we require our dimensionality reduction maps to be \emph{data-oblivious}
since we are constrained by replicability 
requirements.

The cornerstone result in dimensionality reduction for Euclidean distance
is the Johnson-Lindenstrauss lemma (cf. \Cref{thm:jl-lemma}),
which states that there is a distribution over linear maps $\pi_{d, m}$ from $\R^d\to \R^m$
that approximately preserves the norm of any $x\in \R^d$
with constant probability
for some target dimension $m$.
In \citet{makarychev2019performance},
the authors prove \Cref{thm:sample dim reduction},
which roughly states that it suffices to take $m = \tilde O( \nicefrac{p^4}{\varepsilon^2} \log(\nicefrac1\delta) )$ in order to preserve $(k, p)$-clustering costs when the domain is finite.
Firstly,
we extend \Cref{thm:sample dim reduction} to the distributional setting
by implicitly approximating the distribution with a weighted $\varepsilon$-net.
Then,
we implicitly map the $\varepsilon$-net
onto the low-dimensional space and solve the clustering 
problem there.
An important complication we need to overcome
is that this mapping preserves the costs that correspond
to \emph{partitions}\footnote{Roughly speaking, a solution corresponds to a partition when the center of each cluster is its center of mass.} of the data and not arbitrary solutions.
Because of that, it is not clear how we can ``lift'' the
solution from the low-dimensional space to the original
space. Thus, instead of outputting $k$ points that correspond
to the centers of the clusters, our algorithm outputs 
a \emph{clustering function} $f:\mcal X \rightarrow [k]$, which takes as input a point
$x \in \mcal X$ and returns the label of the cluster it belongs to.
The replicability guarantees of the algorithm state that, with probability $1-\rho$,
it will output the \emph{same} function across two executions.
In \Cref{apx:outline dim reduction},
we describe this function.
We emphasize that for each $x \in \mcal X$, the running time of
$f$ is polynomial in $k,d,p,\nicefrac1\varepsilon,\log(\nicefrac1\delta).$
Essentially, this function maps $x$ onto the low-dimensional
space using the same projection map $\pi$ as our algorithm 
and then finds the nearest center of $\pi(x)$ in the low-dimensional space.
For full details,
we refer the reader to \Cref{sec:dimensionality reduction}.
We are now ready to state the result formally.


\begin{restatable}[\Cref{thm:euclidean clustering informal}; Formal]{thm}{kpClusteringEuclidean}\label{thm:euclidean clustering algorithm}
  Let $\varepsilon, \rho\in (0, 1)$ and $\delta\in (0, \nicefrac\rho3)$.
  Given a $\beta$-approximation oracle for weighted Euclidean $k$-medians ($k$-means),
  there is a $\rho$-replicable algorithm that outputs a clustering function
  such that with probability at least $1-\delta$,
  the cost of the partition is at most $(1+\varepsilon)\beta \OPT$.
  Moreover,
  the algorithm has sample complexity
  \begin{align*}
      \tilde O \left( \poly\left( \frac{kd}{\rho \OPT} \right) \left( \frac{2\sqrt m}{\varepsilon} \right)^{O(m)}\log\frac1\delta \right) \,,
  \end{align*}
  where $m = O\left( \frac1{\varepsilon^2} \log \frac{k}{\delta \varepsilon} \right)$.
\end{restatable}

\section{Running Time for \texorpdfstring{$(k, p)$}{(k, p)}-Clustering}
All of our algorithms terminate in $O(\poly(n))$ time
where $n$ denotes the sample complexity.
See \Cref{tab:running time} for more detailed time complexity
of each stage of our approach.
Moreover,
we make $O( \log(\nicefrac\beta{(\varepsilon \OPT)}) )$ calls to the $\beta$-approximation oracle for the combinatorial $(k, p)$-clustering problem within our $\OPT$ estimation subroutine
and then one more call to the oracle in order to output a solution on the coreset.
\section{Replicable \texorpdfstring{$k$}{k}-Centers}\label{sec:k-centers main body}
Due to space limitations, we briefly sketch our 
approach for the $k$-centers problem and kindly refer the reader to \Cref{sec:overview k-centers}. As explained before,
the assumptions we make in this setting state that there exists some ``good'' solution, so that when we draw $n$ i.i.d. samples from $\P$ we observe at least one sample
from each cluster, with constant probability. We first take a fixed grid of side $c$
in order to cover the unit-diameter ball. Then, we sample sufficiently many points
from $\P$. Subsequently, we ``round'' all the points of the sample 
to the centers of the cells of the grid that they fall into
and estimate the probability mass of every cell.
In order to ensure replicability,
we take a random threshold from a predefined interval and discard the points from
all the cells whose mass falls below the threshold. Finally, we call
the approximation oracle using the points that remain. 
Unlike the $(k,p)$-clustering problem
(cf. \Cref{prob:stat cluster}), to the best of our knowledge,
there does not exist
any dimensionality reduction techniques that apply to the $k$-centers
problem. The main result is formally stated in \Cref{thm:formal k-centers sample complexity of reproducible active cells}.

\section{Experiments}
\begin{figure}[H]
  \centering
  \includegraphics[width=\linewidth]{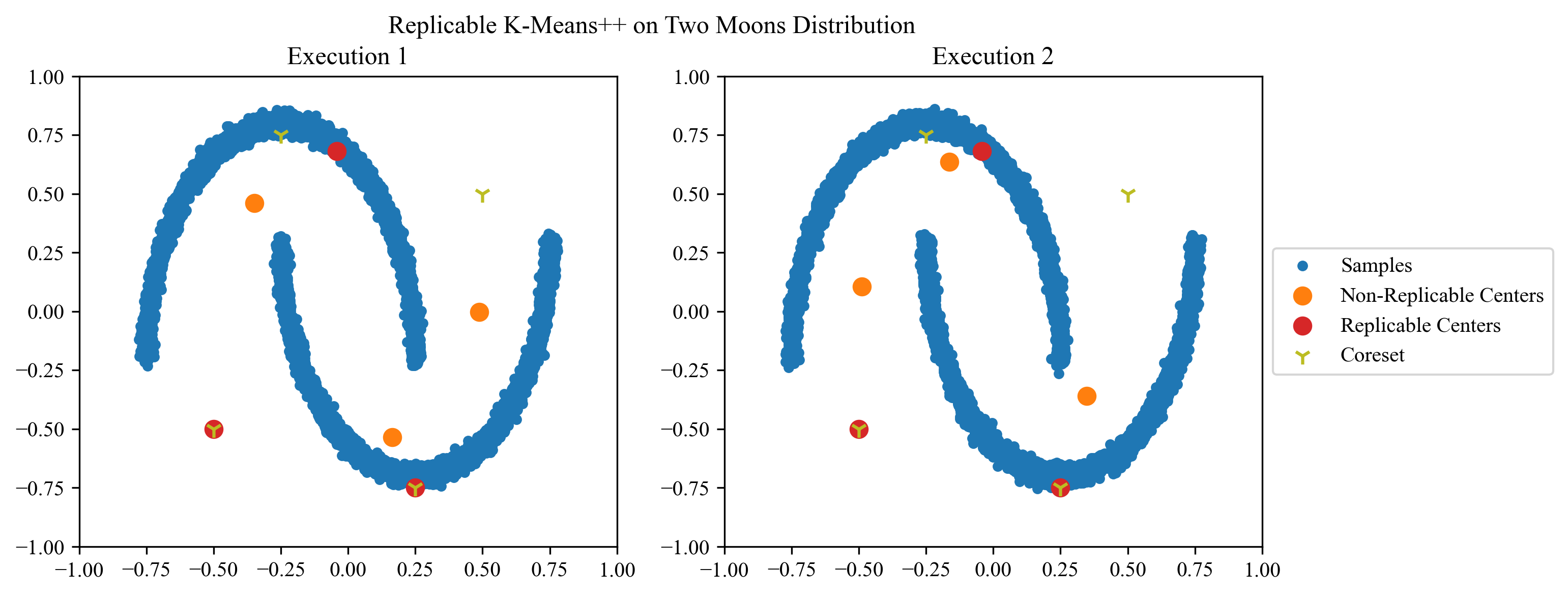}
  \caption{The results of running vanilla vs replicable $k$-Means++ on the two moons distribution for $k=3$.}
  \label{fig:moons}
\end{figure}

\begin{figure}[H]
  \centering
  \includegraphics[width=\linewidth]{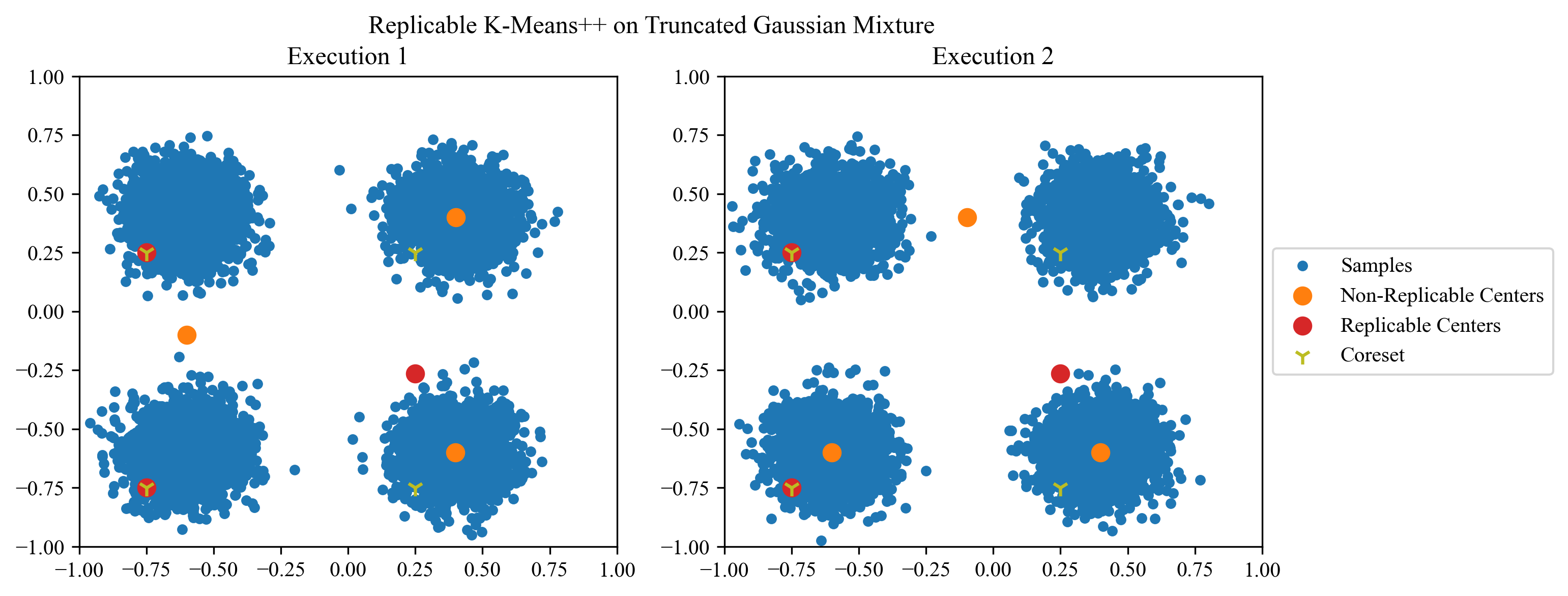}
  \caption{The results of running vanilla vs replicable $k$-Means++ on a mixture of truncated Gaussians distributions for $k=3$.}
  \label{fig:truncnorm}
\end{figure}

We now provide a practical demonstration\footnote{\url{https://anonymous.4open.science/r/replicable_clustering_experiments-E380}} of the replicability of our approach on synthetic data in 2D.
In \Cref{fig:moons},
we leverage the sklearn \citep{scikit-learn} implementation of the popular $k$-means++ algorithm for $k=3$
and compare the output across two executions on the two moons distribution.
In the first experiment,
we do not perform any preprocessing and run $k$-means++ as is,
resulting in different centers across two executions.
In the second experiment,
we compute a replicable coreset for the two moons distribution before running $k$-means++ on the coreset.
This leads to the same centers being outputted across the executions.
Note that the computation for the coreset is performed independently for each execution,
albeit with a shared random seed for the internal randomness.
See also \Cref{fig:truncnorm} for the results of a similar experiment on a mixture of truncated Gaussian distributions.

\section{Conclusion \& Future Work}
In this work,
we designed replicable algorithms  
with strong performance guarantees using
black-box access to approximation oracles for their combinatorial counterparts. 
There are many follow-up research directions that this work can lead to. 
For instance, 
our coreset algorithm adapts the coreset algorithm of \citet{frahling2005coresets}
by viewing their algorithm as a series of heavy hitter estimations that can be made replicable.
it may be possible to interpret more recent approaches for coreset estimation as a series of statistical operations
to be made replicable
in order to get replicable algorithms
in the statistical $(k,p)$-clustering setting \citep{hu2018nearly}.
It would also be interesting to examine the sensitivity of (replicable) clustering algorithms
to the choice of parameters such as the choice of exponent in the cost function or the measure of similarity of the data.
Another relevant direction is to explore sample complexity lower ounds for statistical clustering,
where little is known even in the non-replicable setting.

\begin{ack}
    Amin Karbasi acknowledges funding in direct support of this work from NSF (IIS-1845032), ONR (N00014-19-1-2406), and the AI Institute for Learning-Enabled Optimization at Scale (TILOS). Grigoris Velegkas is supported by TILOS, the Onassis Foundation, and the Bodossaki Foundation. Felix Zhou is supported by TILOS.
\end{ack}

\clearpage
\bibliography{references}
\bibliographystyle{plainnat}

\clearpage
\appendix
\renewcommand{\thesection}{\Alph{section}}

\section{Tables}

We present the running times required for our sub-routines
in order to ensure each stage is $\rho$-replicable
and succeeds with probability at least 0.99.
The overall output is a constant ratio approximation for Euclidean $k$-medians ($k$-means).
\begin{table}[h]
    \caption{Running Time Overview for Euclidean $k$-Medians ($k$-Means)}
    \label{tab:running time}
    \begin{center}
    \begin{small}
    \begin{sc}
    \begin{tabular}{lc}
      \toprule
      Algorithm & Running Time \\
      \midrule
      OPT Estimation & $\tilde O( \poly( \nicefrac{kd}{\OPT\rho} ) )$ \\
      Dimensionality Reduction & $\tilde O( \poly( \nicefrac{d}{\OPT} ) (\nicefrac{k}{\rho})^{O(\log\log (\nicefrac{k}\rho) )} )$ \\
      Coreset & $\tilde O( \poly( \nicefrac{d}{\OPT} ) (\nicefrac{k}{\rho})^{O(\log\log (\nicefrac{k}\rho) )} )$ \\
      Probability Mass Estimation & $\tilde O( \poly( \nicefrac{d}{\OPT} ) (\nicefrac{k}{\rho})^{O(\log\log (\nicefrac{k}\rho) )} )$ \\
      \bottomrule
    \end{tabular}
    \end{sc}
    \end{small}
    \end{center}
\end{table}

\section{Useful Facts}
\begin{prop}[Bretagnolle-Huber-Carol Inequality; \citep{vaart1997weak}]\label{prop:BHC inequality}
  Suppose the random vector $(Z^{(1)}, \dots, Z^{(N)})$ is multinomially distributed
  with parameters $(p^{(1)}, \dots, p^{(N)})$ and $n$.
  Let $\wh p^{(j)} := \frac1n Z^{(j)}$.
  Then
  \[
    \P\set*{\sum_{j=1}^N \abs{\wh p^{(j)} - p^{(j)}} \geq 2\varepsilon}
    \leq 2^N \exp\left( -2\varepsilon^2 n \right).
  \]

  In particular,
  for any $\varepsilon, \rho\in (0, 1)$,
  sampling
  \[
    n \geq \frac{\ln\frac1\rho + N\ln 2}{2\varepsilon^2}
  \]
  points from a finite distribution
  implies that $\sum_{j=1}^N \abs{\wh p^{(j)} - p^{(j)}} < 2\varepsilon$
  with probability at least $1-\rho$.
\end{prop}

\begin{remark}[Weighted $k$-Means/Medians]
We remark that if we have access to a $\beta$-approximation oracle
for unweighted $k$-means/medians, we can implement a weighted one
by considering multiple copies of the points. In particular, 
in our applications, we get weights that are polynomials 
in the parameters of concern
so this will not affect the stated runtime guarantees.
\end{remark}

\section{Uniform Convergence of \texorpdfstring{$(k, p)$}{(k, p)}-Clustering Costs}\label{sec:uniform convergence}
We would like to estimate OPT
for statistical $k$-medians and statistical $k$-means
by solving the combinatorial problem on a sufficiently large sample size.
However,
while the convergence of cost for a particular set of centers is guaranteed by standard arguments, e.g. Chernoff bounds,
we require the stronger statement that convergence holds simultaneously for \emph{all} possible choices of centers.

Similar results can be found \citep{ben2007framework},
with the limitation that centers are either chosen from the sample points,
or chosen as centers of mass of the clusters they induce.
While this suffices to achieve constant ratio approximations,
we would like to choose our centers anywhere in $\mcal B_d$.
One reason for doing so is for replicability
as we have no control over the location of samples from two independent executions so we need
to output centers that are not overly reliant on its specific input.
Another is for dimensionality reduction,
which we will see later.

\subsection{History of Uniform Laws of Large Numbers}
Uniform laws of large numbers generalize convergence results for a finite number of random variables
to possibly uncountable classes of random variables.
The Rademacher complexity and related Gaussian complexity
are canonical techniques in developing these laws
and also have a lengthy history in the study of Banach spaces using probabilistic methods \citep{pisier1999volume,milman1986asymptotic,ledoux1991probability}.
Metric entropy,
along with related notions of expressivity of various function classes,
can be used to control the Rademacher complexity
and are also central objects of study in the field of approximation theory \citep{devore1993constructive,carl1990entropy}.

\subsection{Rademacher Complexity}
Let $\mscr F$ denote a class of functions from $\R^d\to \R$.
For any fixed collection of $n$ points $x_1^n := (x_1, \dots, x_n)$,
we write
\begin{align*}
    f(x_1^n) &:= (f(x_1), \dots, f(x_n)) \\
    \mscr F(x_1^n) &:= \set{f(x_1^n): f\in \mscr F},
\end{align*}
i.e, the restriction of $\mscr F$ onto the sample.
Let $s\in \set{-1, 1}^n$ be a Rademacher random vector.
That is,
$s_i$ takes on values $-1, 1$ with equal probability
independently of other components.
Recall that the \emph{Rademacher complexity} of $\mscr F$ is given by
\[
  \mcal R_n(\mscr F) := \E_{X, s} \left[ \sup_{f\in \mscr F} \frac1n \iprod{s, f(X_1^n)} \right].
\]

\begin{thm}[\citep{wainwright2019high}]\label{thm:rademacher complexity}
  For any $b$-uniformly bounded class of functions $\mscr F$,
  integer $n\geq 1$,
  and error $\varepsilon \geq 0$,
  \[
    \sup_{f\in \mscr F} \abs*{\frac1n \sum_{i=1}^n f(X_i) - \E[f(X)]}
    \leq 2\mcal R_n(\mscr F) + \varepsilon
  \]
  with probability at least $1 - \exp\left( -\frac{n\varepsilon^2}{2b^2} \right)$.

  In particular,
  as long as $\mcal R_n(\mscr F) = o(1)$,
  we have uniform convergence of the sample mean.
\end{thm}

\subsection{Metric Entropy}
We write $B(x, r; \mu)$ to denote the ball of radius $r$ about a point $x$
with respect to a metric $\mu$.
Recall that an \emph{$\varepsilon$-cover} of a set $T$
with respect to a metric $\mu$ is a subset $\theta_1, \dots, \theta_N\sset T$
such that
\[
  T\sset \bigcup_{i\in [N]} B(\theta_i, \varepsilon; \mu).
\]
The \emph{covering number} $N(\varepsilon; T, \mu)$ is the cardinality
of the smallest $\varepsilon$-cover.

\begin{prop}[\citep{wainwright2019high}]\label{prop:covering number}
  Fix $\varepsilon\in (0, 1)$.
  Let $B(R; \norm{\cdot})$ denote the $d$-dimensional ball of radius $R$
  with respect to $\norm{\cdot}$.
  Then
  \[
    N(\varepsilon, B(R; \norm{\cdot}), \norm{\cdot})
    \leq \left( 1 + \frac{2R}\varepsilon \right)^d
    \leq \left( \frac{3R}\varepsilon \right)^d.
  \]
\end{prop}

We say a collection of zero-mean random variables $\set{Y_\theta: \theta\in T}$
is a \emph{sub-Gaussian process}
with respect to a metric $\mu$ on $T$ if
\[
  \E\exp\left[ \lambda(Y_\theta - Y_{\theta'}) \right]
  \leq \exp\left[ \frac{\lambda^2 \mu^2(\theta, \theta')}2 \right]
\]
for all $\theta, \theta'\in T$ and $\lambda\in \R$.

\begin{prop}\label{prop:rademacher subg}
  The canonical Rademacher process
  \[
    Y_\theta := \iprod{s, \theta}
  \]
  is a zero-mean sub-Gaussian process with respect to the Euclidean norm on $T\sset \R^n$.
\end{prop}

\begin{pf}
  Recall that a Rademacher variable is sub-Gaussian with parameter 1.
  Moreover,
  the sum of sub-Gaussian variables is sub-Gaussian with parameter equal to the Euclidean norm of the parameters.
   It follows that $Y_\theta - Y_{\theta'} = \iprod{s, \theta - \theta'}$ is sub-Gaussian with parameter $\norm{\theta - \theta'}_2$.
  The result follows.
\end{pf}

\begin{thm}[One-Step Discretization; \citep{wainwright2019high}]\label{thm:one-step discretization}
  Let $Y_\theta, \theta\in T$ be a zero-mean sub-Gaussian process with respect to the metric $\mu$.
  Fix any $\varepsilon\in [0, D]$
  where $D = \diam(T; \mu)$
  such that $N(\varepsilon; T, \mu) \geq 10$.
  Then
  \begin{align*}
    \E\left[ \sup_{\theta\in T} Y_\theta \right]
    &\leq \E\left[ \sup_{\theta, \theta'\in T} Y_\theta - Y_{\theta'} \right] \\
    &\leq 2\E\left[ \sup_{\theta, \theta'\in T: \mu(\theta, \theta')\leq \varepsilon} Y_\theta - Y_{\theta'} \right] + 4D\sqrt{\log N(\varepsilon; T, \mu)}.
  \end{align*}
\end{thm}
In general,
it may be possible to improve the bound in \Cref{thm:one-step discretization} using more sophisticated techniques
such as Dudley's entropy integral bound \citep{wainwright2019high}.
However,
the simple inequality from \Cref{thm:one-step discretization} suffices for our setting.

\begin{cor}\label{cor:rademacher sup bound}
  Fix any $\varepsilon\in [0, D]$
  where $D = \diam(T; \mu)$
  such that $N(\varepsilon; T, \mu) \geq 10$.
  The canonical Rademacher process $Y_\theta = \iprod{s, \theta}$
  for $\theta\in T\sset \R^n$
  satisfies
  \[
    \E\left[ \sup_\theta Y_\theta \right]
    \leq 2\varepsilon\sqrt n + 4D\sqrt{\log N(\varepsilon; T, \norm{\cdot}_2)}.
  \]
\end{cor}

\begin{pf}
  By \Cref{prop:rademacher subg},
  $Y_\theta$ is a zero-mean sub-Gaussian process
  and we can apply \Cref{thm:one-step discretization} to conclude that
  \begin{align*}
    \E\left[ \sup_{\norm{\theta - \theta'}_2\leq \varepsilon} Y_\theta - Y_\theta' \right]
    &= 2\E\left[ \sup_{\norm{v}_2\leq \varepsilon} \iprod{v, s} \right] \\
    &\leq 2\E \left[ \norm{s}_2\cdot \norm{v}_2 \right] \\
    &\leq 2\varepsilon\sqrt n. \qedhere
  \end{align*}
\end{pf}

\Cref{cor:rademacher sup bound} gives us a way to control the Rademacher complexity of a function class
whose co-domain is well-behaved.
We make this notion rigorous in the next section.

\subsection{Uniform Convergence of \texorpdfstring{$(k, p)$}{(k, p)}-Clustering Cost}
For a fixed set of centers $F$,
let $\kappa_F^p: \R^d\to \R$ be given by
\[
  \kappa_F^p(x) := \kappa(x, F)^p.
\]
Define $\mcal F := \set{F\sset \mcal B_d: \card F = k}$.
We take our function class to be
\[
  \mscr F := \set{\kappa_F^p: F\in \mcal F}.
\]

Let $\mu: \mcal F\times\mcal F\to \R$ be given by
\[
  \mu(A, B) := \max_{a\in A, b\in B} \kappa(a, b).
\]
\begin{prop}
  $\mu$ is a metric on $\mcal F$.
\end{prop}

\begin{pf}
  It is clear that $\mu$ is symmetric and positive definite.
  We need only show that the triangle inequality holds.
  Fix $A, B, C\in \mcal F$
  and suppose $a\in A, c\in C$ are such that $\kappa(a, c) = \mu(A, C)$.
  Then
  \begin{align*}
    \mu(A, C)
    &= \kappa(a, c) \\
    &\leq \kappa(a, b) + \kappa(b, c) &&\forall b\in B \\
    &\leq \mu(A, B) + \mu(B, C)
  \end{align*}
  as desired.
\end{pf}

\begin{prop}
  $\mscr F$ is $p$-Lipschitz parameterized with respect to the metric $\mu$.
\end{prop}

\begin{pf}
  We have
  \begin{align*}
    \abs{\kappa_F^1(x) - \kappa_{F'}^1(x)}
    &= \abs{\kappa(x, F(x)) - \kappa(x, F'(x))} \\
    &\leq \abs{\kappa(x, F'(x)) + \kappa(F'(x), F(x)) - \kappa(x, F'(x))} \\
    &= \kappa(F'(x), F(x)) \\
    &\leq \mu(F, F').
  \end{align*}

  Now,
  the function $g(x): \left[ 0, 1 \right]\to \R$ given by $x\mapsto x^p$
  is $p$-Lipschitz by the mean value theorem:
  \[
    \abs{g(x) - g(y)} \leq \sup_{\xi\in [0, 1]} g'(\xi) \abs{x-y} \leq p \abs{x-y}.
  \]
  
  The result follows by the fact that
  the composition of Lipschitz functions is Lipschitz
  with a constant equal to the product of constants from the composed functions.
\end{pf}

\begin{thm}\label{thm:rademacher centers bound}
  For any $\varepsilon\in \left[ 0, D=\sqrt{n} \right]$
  such that $N(\varepsilon; \mscr F(x_1^n), \norm{\cdot}_2) \geq 10$,
  \[
    \mcal R_n(\mscr F)
    \leq \frac{2\varepsilon}{\sqrt n} + 4\sqrt{\frac{kd \log \frac{3\sqrt{n} p}{\varepsilon}}{n}}
    = o(1).
  \]
\end{thm}

\begin{pf}
  Remark that $\kappa_F^p(x_1^n)$ is $\sqrt{n} p$-Lipschitz with respect to the metric $\mu$.
  Indeed,
  \[
    \sum_{i=1}^n \abs{\kappa_F^p(x_i) - \kappa_{F'}^p(x_i)}^2 \leq n p^2\mu(F, F')^2.
  \]

  Now,
  $D := \diam(\mscr F(x_1^n); \norm{\cdot}_2) = \sqrt{n}$.
  We can apply \Cref{cor:rademacher sup bound}
  with $T = \mscr F(x_1^n)$
  to see that
  \begin{align*}
    \mcal R_n(\mscr F)
    &= \frac1n \E\left[ \sup_\theta Y_\theta \right] \\
    &\leq \frac{2\varepsilon}{\sqrt n} + 4\sqrt{\frac{\log N(\varepsilon; \mscr F(x_1^n), \norm{\cdot}_2)}{n}}.
  \end{align*}

  Now,
  since $\kappa_F^p(x_1^n)$ is $\sqrt{n} p$-Lipschitz,
  a $\frac{\varepsilon}{\sqrt{n} p}$-cover for $\mcal F$
  yields an $\varepsilon$-cover of $\mscr F(x_1^n)$.
  Hence
  \[
    N(\varepsilon; \mscr F(x_1^n), \norm{\cdot}_2)
    \leq N\left( \frac\varepsilon{\sqrt{n} p}; \mcal F, \mu \right).
  \]
  Note that the cross product of $k$ $\varepsilon$-covers of $\mcal B$
  is an $\varepsilon$-cover of $\mcal F$.
  Hence
  \begin{align*}
    N\left( \frac\varepsilon{\sqrt{n} p}; \mcal F, \mu \right)
    &\leq N\left( \frac\varepsilon{\sqrt{n} p}; \mcal B_d, \kappa \right)^k \\
    &\leq \left( \frac{3 \sqrt{n} p}{\varepsilon} \right)^{kd}. &&\text{\Cref{prop:covering number}}
  \end{align*}

  Substituting this bound on the covering number of $\mscr F(x_1^n)$ concludes the proof.
\end{pf}
Note that the Lipschitz property was crucial to ensure that the exponent in the covering number does not contain $n$.

\begin{thm}\label{thm:uniform convergence}
  Fix $\varepsilon, \delta\in (0, 1)$.
  Then with
  \[
    O\left( \frac{k^2 d^2}{\varepsilon^4} \log\frac{p}\delta \right)
  \]
  i.i.d. samples from $\P$,
  \[
    \abs*{\wh \cost(F) - \cost(F)} \leq \varepsilon
  \]
  with probability at least $1-\delta$
  for any set of centers $F\sset \mcal B_d$.
\end{thm}

\begin{pf}
  Choose $\varepsilon := 3\leq D=\sqrt n$.
  By \Cref{thm:rademacher centers bound},
  \begin{align*}
    &\mcal R_n(\mscr F) \\
    &\leq \frac{2\cdot 3}{\sqrt n} + 4 \sqrt{\frac{kd \log \frac{3\sqrt{n} p}{3}}{n}} \\
    &\leq \frac{6}{\sqrt n} + 4 \sqrt{\frac{kd \log \sqrt{n} p}{n}} \\
    &\leq \frac{6}{\sqrt n} + 4 \sqrt{\frac{kd \log p}{n}} + 4 \sqrt{\frac{kd \log \sqrt n}{n}} &&\sqrt{a+b}\leq \sqrt a + \sqrt b \\
    &\leq \frac{6}{\sqrt n} + 4 \sqrt{\frac{kd \log p}{n}} + 4 \sqrt{\frac{kd}{\sqrt n}}. &&\log \sqrt n\leq \sqrt n
  \end{align*}
  
  Fix some $\varepsilon_1\in (0, 1)$.
  We have
  \begin{align*}
    &\frac{6}{\sqrt n} \leq \varepsilon_1 \\
    &\iff n\geq \frac{36}{\varepsilon_1^2}.
  \end{align*}
  Similarly,
  we have
  \begin{align*}
    &4\sqrt{\frac{kd \log p}{n}} \leq \varepsilon_1 \\
    &\iff n\geq \frac{16 kd \log p}{\varepsilon_1^2}.
  \end{align*}
  Finally,
  \begin{align*}
    &4 \sqrt{\frac{kd}{\sqrt n}} \leq \varepsilon_1 \\
    &\iff n\geq \frac{256k^2 d^2}{\varepsilon_1^4}.
  \end{align*}
  By taking the maximum of the three lower bounds,
  we conclude that
  \[
    n\geq \frac{256 k^2 d^2}{\varepsilon_1^4}\log p
    \implies \mcal R_n(\mscr F) \leq 3\varepsilon_1.
  \]
  
  Fix $\delta\in (0, 1)$.
  Observe that $\mscr F$ is $1$-uniformly bounded.
  Thus by \Cref{thm:rademacher complexity},
  we require 
  \[
    \max\left( \frac{2}{\varepsilon_1^2} \log\frac1\delta, \frac{256 k^2 d^2}{\varepsilon_1^4}\log p \right)
    \leq \frac{256 k^2 d^2}{\varepsilon_1^4} \log\frac{p}\delta
  \]
  samples in order to guarantee that
  \begin{align*}
    \sup_{\kappa_F^p\in \mscr F} \abs*{\frac1n \sum_{i=1}^n \kappa_F^p(X_i) - \E \kappa_F^p(X)}
    &\leq 2\mcal R_n(\mscr F) + \varepsilon_1 \\
    &\leq 7\varepsilon_1
  \end{align*}
  with probability at least $1-\delta$.

  Choosing $\varepsilon_1 = \frac{\varepsilon}7$ concludes the proof.
\end{pf}

\section{\texorpdfstring{$(k, p)$}{(k, p)}-Clustering}
In this section,
we provide the full proofs for \Cref{thm:statistical k-medians algorithm}
and \Cref{thm:statistical k-means algorithm},
which state the guarantees for the replicable $k$-medians
and $k$-means algorithms,
respectively.
First,
we describe a useful subroutine for replicable heavy hitters estimation in \Cref{sec:heavy-hitters}.
This subroutine is crucial to the replicable coreset algorithm (cf. \Cref{alg:reproducible coreset}),
which we analyze in \Cref{sec:reproducible coreset appendix}.
Once we have a coreset,
it remains to solve the statistical $(k, p)$-clustering problem on a finite distribution.
We describe a replicable algorithm for this in \Cref{sec:multinomial estimation}.
Our coreset algorithm assumes the knowledge of some constant ratio estimate of OPT.
In \Cref{sec:k-medians OPT estimation},
we show how to output such an estimate replicably.
Finally,
we summarize our findings in \Cref{sec:statistical clustering summary}.

\subsection{Warm-Up: Replicable SQ Oracle and \texorpdfstring{$\varepsilon$}{Epsilon}-Covers}
In order to give some intuition to the reader,
we first show how we can use a subroutine that was developed in 
\citet{impagliazzo2022reproducibility} in order to derive some results
in the setting we are studying.
We first need to define the \emph{statistical query} model that was introduced in 
\citet{kearns1998efficient}

\begin{defn}[Statistical Query Oracle; \citep{kearns1998efficient}]\label{def:Statistical Query Oracle} 
    Let $\mathcal{D}$ be a distribution over the domain $\mathcal{X}$
    and $\phi: \mathcal{X}^n \to \R$ be a statistical query
    with true value
    \[
        v^\star := \lim_{n\to \infty} \phi(X_1, \dots, X_n)\in \R.
    \]
    Here $X_i\sim_{i.i.d.} \mcal D$
    and the convergence is understood in probability or distribution.
    Let $\varepsilon,\delta \in (0,1)^2$.
    A \emph{statistical query (SQ) oracle} outputs 
    a value $v$ such that $\abs*{v - v^\star} \leq \varepsilon$
    with probability at least $1-\delta$.
\end{defn}

The simplest example of a statistical query is the sample mean
\[
    \phi(X_1, \dots, X_n)
    = \frac1n \sum_{i=1}^n X_i.
\]

The idea is that by using a sufficiently large number of samples,
an SQ oracle returns an estimate of the expected value of a statistical query whose range is bounded.
\citet{impagliazzo2022reproducibility} provide a replicable implementation of an SQ oracle
with a mild blow-up in the sample complexity
which we state below.

\begin{thm}[Replicable SQ Oracle; \citep{impagliazzo2022reproducibility}]
\label{thm:reproducible sq oracle}
    Let $\varepsilon, \rho\in (0, 1)$ and $\delta\in (0, \nicefrac\rho3)$.
  Suppose $\phi$ is a statistical query with co-domain $[0, 1]$.
  There is a $\rho$-replicable SQ oracle to estimate its true value
  with tolerance $\varepsilon$ and failure rate $\delta$.
  Moreover,
  the oracle has sample complexity
  \[
    \tilde O\left( \frac1{\varepsilon^2 \rho^2} \log\frac1\delta \right).
  \]
\end{thm}
The interpretation of the previous theorem is that we can replicably estimate
statistical queries whose range is bounded. 

We now explain how we can use this result for the statistical $k$-means problem under the Euclidean metric,
but it is not hard to extend the approach to more general $(k,p)$-clustering problems.
Consider a \emph{fixed} set of centers $F.$ Then, we can replicably estimate the cost of this solution
using \Cref{thm:reproducible sq oracle} within an additive accuracy $\varepsilon$ 
and confidence $\delta$ using $O\left( \varepsilon^{-2}\rho^{-2} \log (\nicefrac1\delta) \right)$ samples.
Thus,
a natural approach is to consider an $\varepsilon$-cover of the unit-diameter ball
and then exhaustively search among solutions whose centers coincide with elements of the $\varepsilon$-cover.
This is outlined in \Cref{alg:reproducible k-means with fixed grid}.
We are now ready to state our results.

\begin{algorithm}[h]
\caption{Replicable $k$-Means with $\varepsilon$-Cover}\label{alg:reproducible k-means with fixed grid}
\begin{algorithmic}
  \STATE {\bfseries rKMeansCover:} {distribution $\P$, error $\varepsilon$, replicability $\rho$, confidence $\delta$}
  \STATE $G \gets \nicefrac\varepsilon3$-cover over the $d$-dimensional unit-diameter ball
  \STATE $\mcal F \gets \{F \subseteq G: |F| = k \}$ \COMMENT{We restrict the solutions to the $\varepsilon$-cover.}
  \STATE Output $\argmin_{F\in \mcal F} \operatorname{rSQ-Oracle}(\cost(F), \nicefrac\varepsilon3, \nicefrac\rho{\card{\mcal F}}, \nicefrac\delta{\card{\mcal F}})$
\end{algorithmic}
\end{algorithm}

\begin{lem}\label{lem:k-means with fixed grid}
For any $\varepsilon, \delta, \rho \in (0,1), \delta < \nicefrac\rho3$,
\Cref{alg:reproducible k-means with fixed grid} is $\rho$-replicable
and outputs a solution $F$ whose cost is at most $\OPT + \varepsilon$
with probability $1-\delta$.
Moreover, it has sample complexity
\[
    \tilde O\left( \frac{(\nicefrac9\varepsilon)^{2kd}}{\rho^2\varepsilon^2} \log \frac1\delta \right) \,.
\]
\end{lem}

\begin{pf}[\Cref{lem:k-means with fixed grid}]
    We first argue about the replicability of \Cref{alg:reproducible k-means with fixed grid}.
    Since we make $\card{\mcal F}$ calls to the replicable SQ subroutine with parameter $\nicefrac\rho{\card{\mcal F}}$,
    the overall replicability of the algorithm follows by taking a union bound.

    Let us now focus on the correctness of the algorithm.
    Let $F^*$ be the optimal solution.
    Consider the solution that we get when we move the centers of $F^*$
    to the closest point of $G$
    and let us denote it by $\hat{F}^*$. 
    Notice that the cost of $\hat{F}^*$ is at most $\OPT + \nicefrac\varepsilon3$.
    Furthermore,
    by a union bound,
    all the calls to the SQ oracle will return an estimate that is within an additive $\nicefrac\varepsilon3$-error of the true cost.
    This happens with probability
    at least $1-\delta$
    and we condition on this event for the rest of the proof.
    Thus the estimated cost of the solution
    $\hat{F}^*$ will be at most $\OPT + \nicefrac{2\varepsilon}3$.
    Let $\widetilde{F}$ be the solution that we output.
    Its estimated cost is at most that of $\hat{F}^*$
    and so its true cost will be at most $\OPT + \varepsilon.$
    This concludes the proof of correctness.

    Lastly,
    we argue about the sample complexity of the algorithm.
    By \Cref{prop:covering number},
    \[
        \card{\mcal F} \leq O\left((\nicefrac9\varepsilon)^{kd}\right).
    \]
    By plugging this value into the sample complexity from \Cref{thm:reproducible sq oracle},
    we get
    \[
        O\left( \frac{(\nicefrac9\varepsilon)^{2kd}}{\rho^2\varepsilon^2} \log \frac{(\nicefrac9\varepsilon)^{kd}}{\delta} \right) \,. \qedhere
    \]
\end{pf}

Although \Cref{alg:reproducible k-means with fixed grid} provides some basic guarantees, there are several caveats with this approach.
Namely,
we can only get additive approximations
and the dependence of the sample and time complexity on both $k, d$ is exponential.

In the following sections,
we will explain how we can overcome these issues.
As we alluded to before, our approach combines ideas from coresets estimation through hierarchical 
grids \citep{frahling2005coresets}, uniform convergence through metric entropy
\citep{wainwright2019high}, and dimensionality reduction techniques \citep{makarychev2019performance}.

\subsection{Replicable Heavy-Hitters}\label{sec:heavy-hitters}
In this section,
we present a replicable heavy hitters algorithm
which is inspired by \citet{impagliazzo2022reproducibility}
and has an improved sample complexity by a factor of $O(\nicefrac1{(v-\varepsilon)}).$ We
believe that this result could be of independent interest since the 
replicable heavy hitters algorithm has many applications as a subroutine in more
complicated algorithms \citep{impagliazzo2022reproducibility}. Intuitively,
this algorithm consists of two phases.
In the first phase,
we estimate all the candidate
heavy hitters and reduce the size of the domain.
In the second phase,
we estimate the mass of these candidates. We present a new analysis of the second
phase using \Cref{prop:BHC inequality}.

\begin{algorithm}[h]
\caption{Replicable Heavy Hitters}\label{alg:reproducible heavy hitters}
\begin{algorithmic}
  \STATE {\bfseries rHeavyHitters:}{distribution $\P$, target $v$, error $\varepsilon$, replicability $\rho$, confidence $\delta$}
  \IF {$\card{\mcal X} < \frac{\ln\frac2{\delta (v-\varepsilon)}}{v-\varepsilon}$}
    \STATE $\hat{\mcal X} \gets \mcal X$
  \ELSE
    \STATE $\hat{\mcal X} \gets \set*{\text{$\frac{\ln\frac2{\delta (v-\varepsilon)}}{v-\varepsilon}$ samples $\sim \P$}}$
  \ENDIF
  \STATE $S \gets \set*{\text{$\frac{648\ln\nicefrac2\delta + 648\left( \card{\hat{\mcal X}} + 1 \right)\ln 2}{\rho^2\varepsilon^2}$ samples $\sim \P$}}$
  \STATE Choose $v'\in \left[ v-\nicefrac23\varepsilon, v-\nicefrac13\varepsilon \right]$ uniformly randomly
  \STATE Output all $x\in \hat{\mcal X}$ such that $\hat \P_S(x) \geq v'$ \COMMENT{empirical distribution}
\end{algorithmic}
\end{algorithm}

\begin{thm}\label{thm:heavy hitters}
  Fix $v, \rho\in (0, 1)$ and $\varepsilon\in (0, v), \delta\in (0, \nicefrac\rho3)$.
  Then \Cref{alg:reproducible heavy hitters} is $\rho$-replicable
  and returns a list $L$ of elements $x$ such that
  with probability at least $1-\delta$:
  \begin{enumerate}[(a)]
    \item If $\P(x) \leq v-\varepsilon$,
      then $x\notin L$.
    \item If $\P(x) \geq v$,
      then $x\in L$.
  \end{enumerate}
  Moreover,
  it has sample complexity
  \[
    \tilde O\left( \min\left( \card{\mcal X},
    \frac1{(v-\varepsilon)} \right)
    \frac1{\rho^2 \varepsilon^2} \log\frac1\delta \right).
  \]
\end{thm}

\begin{pf}
  First,
  we wish to capture all $v-\varepsilon$ heavy hitters in $\hat{\mcal X}$.
  If $\mcal X$ is sufficiently small,
  this is easy.
  Otherwise,
  we fail to observe each $(v-\varepsilon)$ heavy hitter
  with probability at most $(1-v+\varepsilon)^{\card{\hat{\mcal X}}}$.
  By a union bound over all $(v-\varepsilon)^{-1}$ possible heavy hitters,
  we fail to capture all $v-\varepsilon$ heavy hitters with probability at most
  \begin{align*}
    \frac{(1-v+\varepsilon)^{\card{\hat{\mcal X}}}}{v-\varepsilon}
    &\leq \frac1{v-\varepsilon} \exp\left[ -(v-\varepsilon)\cdot \frac{\ln\frac2{\delta (v-\varepsilon)}}{v-\varepsilon} \right] \\
    &\leq \frac\delta2.
  \end{align*}
  Moving forward,
  we condition on this step succeeding.

  Next,
  consider $\hat p_1, \hat p_2$,
  the mass estimates over the course of two runs
  supported on the candidate sets $\hat{\mcal X}_1, \hat{\mcal X}_2$.
  both follow (possibly different) multinomial distributions
  of dimension at most $\card{\hat{\mcal X}} + 1$
  with unknown mean parameters $p_1(x)$ for $x\in \hat{\mcal X}_1\cup \set{y_1}$,
  $p_2(x)$ for $x\in \hat{\mcal X}_2\cup \set{y_2}$,
  respectively,
  and $\card S$.
  Here $y_1, y_2$ are dummy elements for observations beyond $\hat{\mcal X}_1, \hat{\mcal X}_2$.
  Suppose we draw $\frac{\ln\nicefrac2\delta + \left( \card{\hat{\mcal X}} + 1 \right)\ln 2}{2\varepsilon'^2}$ samples over each of two runs
  to yield estimates $\hat p_1, \hat p_2$.
  By \Cref{prop:BHC inequality},
  \begin{align*}
    \sum_{x\in \hat{\mcal X}_1\cup \set{y_1}} \abs{\hat p_1(x) - p_1(x)} &< 2\varepsilon' \\
    \sum_{x\in \hat{\mcal X}_2\cup \set{y_2}} \abs{\hat p_2(x) - p_2(x)} &< 2\varepsilon'
  \end{align*}
  each with probability at least $1-\nicefrac{\delta}2$.
  Moving forward,
  we condition on this step succeeding.

  Now, 
  consider choosing $v'\in [v-\nicefrac{2\varepsilon}3, v-\nicefrac\varepsilon3]$ uniformly at random.

  \underline{Correctness:}
  By choosing $2\varepsilon'<\nicefrac\varepsilon3$,
  any element $x$ with true mass at least $v$
  will have empirical mass strictly more than $v-\nicefrac\varepsilon3\geq v'$.
  Similarly,
  any element $x$ with true mass at most $v-\varepsilon$
  will have empirical mass strictly less than $v-\nicefrac{2\varepsilon}3\leq v'$.
  Thus we satisfy the correctness guarantees.
  Note that we satisfy this guarantee with probability at least $1-\delta$.

  \underline{Replicability:}
  $v'$ lands in between some $\hat p_1(x), \hat p_2(x)$
  for $x\in \hat{\mcal X}_1\cap \hat{\mcal X}_2$
  with total probability at most
  \[
    \frac{\sum_{x\in \hat{\mcal X}_1\cap \hat{\mcal X}_2} \abs{\hat p_1(x) - \hat p_2(x)}}{\varepsilon/3}
    \leq \frac{\sum_{x\in \hat{\mcal X}_1\cap \hat{\mcal X}_2} \abs{\hat p_1(x) - p_1(x)}}{\varepsilon/3}
    + \frac{\sum_{x\in \hat{\mcal X}_1\cap \hat{\mcal X}_2} \abs{\hat p_2(x) - p_2(x)}}{\varepsilon/3}
  \]
  In addition,
  we wish for $v'$ to avoid landing ``below'' any $\hat p_1(x), \hat p_2(x)$
  where $x\notin \hat{\mcal X}_1\cap \hat{\mcal X}_2$.
  This happens with probability at most
  \[
    \frac{\sum_{x\in \hat{\mcal X}_1\cup \set{y_1}\setminus \hat{\mcal X}_2}
    \abs{\hat p_1(x) - p_1(x)}}{\varepsilon/3}
    + \frac{\sum_{x\in \hat{\mcal X}_2\cup \set{y_2}\setminus \hat{\mcal X}_1}
    \abs{\hat p_2(x) - p_2(x)}}{\varepsilon/3}.
  \]
  The sum of the two expressions above is at most $\frac{4\varepsilon'}{\varepsilon/3}$.
  Thus we set
  \begin{align*}
    \frac{4\varepsilon'}{\varepsilon/3} &\leq \frac{\rho}3 \\
    \varepsilon' &\leq \frac{\rho \varepsilon}{36}.
  \end{align*}

  The probability of failing to be replicable is then at most $1 - \nicefrac\rho3 - 2\delta \geq 1-\rho$.

  \underline{Sample Complexity:}
  Recall that we set 
  \[
    \varepsilon'
    = \min\left( \frac{\rho \varepsilon}{36}, \frac{\varepsilon}6 \right)
    = \frac{\rho \varepsilon}{36}.
  \]

  If $\card{\mcal X}$ is sufficiently small,
  the sample complexity becomes
  \[
    \frac{\ln\frac2\delta + \left( \card{\mcal X} + 1 \right)\ln 2}{2\varepsilon'^2}
    = \frac{648 \ln\frac2\delta + 648 \left( \card{\mcal X} + 1 \right)\ln 2}{\rho^2 \varepsilon^2}
  \]
  Otherwise,
  the sample complexity is
  \[
    \frac{\ln \frac2{\delta(v-\varepsilon)}}{v-\varepsilon}
    + \frac{648\ln\frac2\delta}{\rho^2\varepsilon^2}
    + \frac{648 \left( \ln\frac2{\delta(v-\varepsilon)} + 1 \right) \ln 2}{\rho^2\varepsilon^2(v-\varepsilon)}.
  \]
\end{pf}

\subsection{Replicable Coreset}\label{sec:reproducible coreset appendix}
We now prove that \Cref{alg:replicable quad tree} replicably yields an $\varepsilon$-coreset.
In \Cref{sec:reproducible coreset pseudocode},
we state \Cref{alg:reproducible coreset},
an equivalent description of \Cref{alg:replicable quad tree}.
Then,
we show that it is indeed replicable in \Cref{sec:coreset reproducibility}.
Following this,
we prove that the cost of any set of centers $F\sset \mcal B_d$
is preserved up to an $\varepsilon$-multiplicative error in \Cref{sec:expected shift}.
In \Cref{sec:coreset size},
we show that the outputted coreset is of modest size.
Last but not least,
we summarize our findings in \Cref{sec:coreset summary}.

\subsubsection{Pseudocode}\label{sec:reproducible coreset pseudocode}
Before presenting the analysis,
we state an equivalent algorithm to \Cref{alg:replicable quad tree}
which uses precise notation that is more suitable for analysis.

\begin{algorithm}[h]
\caption{Replicable Coreset; \Cref{alg:replicable quad tree} Equivalent}\label{alg:reproducible coreset}
\begin{algorithmic}[1]
  \STATE {\bfseries rCoreset}(distribution $\P$, accuracy $\varepsilon$, exponent $p$, replicability $\rho$, confidence $\delta$):
  \STATE Init $\mcal H[i] \gets \varnothing$ for all $i$
  \STATE Init $\mcal L[i] \gets \varnothing$ for all $i$
  \STATE Init $\mcal S[i] \gets \varnothing$ for all $i$
  \STATE
  \STATE $\mcal H(0) \gets \set*{\left[ -\nicefrac12, \nicefrac12 \right]^d}$
  \FOR {$i\gets 1; H[i-1]\neq \varnothing; i\gets i+1$}
  \IF {$(2^{-i+1} \Delta)^p \leq \nicefrac{\varepsilon \Lambda}{5}$}
      \STATE $\mcal S[i] \gets \children(\mcal H[i-1])$
    \ELSE
    \STATE $\mcal H[i] \gets \children(\mcal H[i-1]) \cap \rHeavyHitters\left( \P_i, v=\frac{\gamma\cdot \Lambda}{2^{-pi}}, \nicefrac{v}2, \nicefrac\rho{t}, \nicefrac\delta{t} \right)$ \COMMENT{$t$ is an upper bound on the number of layers.}
      \STATE $\mcal L[i] \gets \children(\mcal H[i-1])\setminus \mcal H[i]$
    \ENDIF
  \ENDFOR
  \STATE
  \FOR {$j\gets i-1; j\geq 0; j\gets j-1$}
    \FOR {$Z\in \mcal H[j]$}
      \IF {$\children(Z)\cap \mcal H[j+1] = \varnothing$}
        \STATE $R'(Z) \gets \rChoice(Z)$
      \ELSE
        \STATE $R'(Z) \gets R'(\rChoice(\children(Z)\cap \mcal H[j+1]))$
      \ENDIF
    \ENDFOR
    \FOR {$Z\in \mcal L[j+1]$}
      \STATE $R(Z) \gets R'(\parent(Z))$
    \ENDFOR
    \FOR {$Z\in \mcal S[j+1]$}
      \STATE $R(Z) \gets R'(\parent(Z))$
    \ENDFOR
  \ENDFOR
  \STATE
  \STATE Output $R$
\end{algorithmic}
\end{algorithm}

Starting with the grid that consists of a single cell $[-\nicefrac12, \nicefrac12]^d$ at layer $0$,
we recursively subdivide each cell into $2^{d}$ smaller cells of length $2^{-i}$ at layer $i$.
Each subdivided cell is a \emph{child} of the larger cell,
the containing cell is the \emph{parent} of a child cell,
and cells sharing the same parent are \emph{siblings}.
Let $\mcal G_i$ denote the union of cells at the $i$-th level.
We write $\P_i$ to denote the discretized distribution to $\mcal G_i$.
In other words,
$\P_i = \eval{\P}_{\sigma(\mcal G_i)}$ is the restriction of $\P$
to the smallest $\sigma$-algebra containing $\mcal G_i$.
We say a cell on the $i$-th level is \emph{heavy}
if the replicable heavy hitters algorithm (cf. \Cref{alg:reproducible heavy hitters}) returns it
with input distribution $\P_i$,
heavy hitter threshold $v = \frac{\gamma\cdot \Lambda}{2^{-pi}}$,
and error $\nicefrac v2$
for some parameter $\gamma$ to be determined later.
A cell that is not heavy is \emph{light}.
The recursion terminates either when there are no more heavy cells on the previous layer,
or the current grid length is a sufficiently small fraction of $\OPT$.
In the second case,
we say the last layer consists of \emph{special} cells.

If no child cell of a heavy parent is heavy,
we mark the parent.
Otherwise,
we recurse on its heavy children
so that one of its descendants will eventually be marked.
The recursion must stop,
either when there are no more heavy children
or when the side length is no more than a fraction of $\Lambda$.
Note that the light cells with heavy parents
and special cells partition $[-\nicefrac12, \nicefrac12]^d$.

Then,
we build $R$ in reverse as follows.
For a marked heavy cell $Z$,
we set $R(x)$ to be an arbitrary (but replicable) point in $Z$,
e.g. its center,
for all $x\in Z$.
This covers all light cells with no heavy siblings
as well as special cells.
Otherwise,
a cell $Z$ is light with heavy siblings,
and its parent is heavy but unmarked.
Thus, the parent has some marked heavy descendent $Z'$,
and we choose $R(x) = R(x')$
for all $x\in Z$ and $x'\in Z'$.

\subsubsection{Replicability}\label{sec:coreset reproducibility}
\begin{prop}
  \Cref{alg:reproducible coreset} terminates after at most
  \[
    t := \ceil*{\frac1p \log \frac{10\beta \Delta^p}{\varepsilon\cdot \OPT} + 1}
  \]
  layers.
\end{prop}

\begin{pf}
  We have
  \begin{align*}
    &(2^{-i+1} \Delta)^p \leq \frac{\varepsilon \Lambda}{5} \\
    &\iff 2^{p(-i+1)} \leq \frac{\varepsilon \Lambda}{5 \Delta^p} \\
    &\iff p(-i+1) \leq \log \frac{\varepsilon \Lambda}{5 \Delta^p} \\
    &\iff i\geq \frac1p \log \frac{5 \Delta^p}{\varepsilon\cdot \Lambda} + 1.
  \end{align*}
  But $\nicefrac1\Lambda\leq \nicefrac{\beta( 1+\varepsilon )}{\OPT}$,
  concluding the proof.
\end{pf}

\begin{cor}\label{cor:coreset sample complexity}
  \Cref{alg:reproducible coreset} is $\rho$-replicable
  and succeeds with probability at least $1-\delta$.
  Moreover,
  it has sample complexity
  \[
    \tilde O \left( \frac{t^2}{\rho^2\gamma^3\cdot \OPT^3}\log\frac1\delta \right).
  \]
\end{cor}

\begin{pf}
  The only non-trivial random components of \Cref{alg:reproducible coreset} is the heavy hitter estimation.
  By \Cref{thm:heavy hitters},
  the heavy hitters subroutine call on the $i$-th layer
  is $\nicefrac\rho{t}$-replicable
  and succeeds with probability at least $1-\nicefrac\delta{t}$.
  Also,
  it has sample complexity
  \[
    \tilde O \left( \frac{t^2\cdot 2^{3(-pi+1)}}{\rho^2\gamma^3\cdot \Lambda^3}\log\frac1\delta \right).
  \]
  Now,
  \begin{align*}
    \sum_{i=0}^t 2^{-3pi}
    &\leq \sum_{i=0}^\infty 2^{-3pi} \\
    &= O(1).
  \end{align*}

  Thus all in all,
  \Cref{alg:reproducible coreset} is $\rho$-replicable
  and succeeds with probability at least $1-\delta$.
  It also has sample complexity
  \[
    \tilde O \left( \frac{t^2}{\rho^2\gamma^3\cdot \Lambda^3}\log\frac1\delta \right).
  \]
  Since $\nicefrac1\Lambda\leq \nicefrac{\beta( 1+\varepsilon )}{\OPT}$,
  we conclude the proof.
\end{pf}

\subsubsection{Expected Shift}\label{sec:expected shift}
\begin{prop}\label{prop:max shift}
  For any $x\in Z\in \mcal L[i]\cup \mcal S[i]$ outputted by \Cref{alg:reproducible coreset},
  \[
    \kappa(x, R(x)) \leq (2^{-i+1} \Delta).
  \]
\end{prop}

\begin{pf}
  $x$ is represented by some point within its parent.
\end{pf}

\begin{prop}
  The output $\mcal S[t]$ of \Cref{alg:reproducible coreset} satisfies
  \[
    \int_{\mcal S[t]} \kappa(x, R(x))^p d\P(x)
    \leq \frac{\varepsilon \Lambda}{5}.
  \]
\end{prop}

\begin{pf}
  By computation,
  \begin{align*}
    \int_{\mcal S[t]} \kappa(x, R(x))^p d\P(x)
    &\leq \int_{\mcal S[t]} (2^{-i+1} \Delta)^p d\P(x) &&\text{\Cref{prop:max shift}} \\
    &\leq \int_{\mcal S[t]} \frac{\varepsilon \Lambda}{5} d\P(x) &&\text{definition of $t$} \\
    &\leq \int_{\mcal X} \frac{\varepsilon \Lambda}{5} d\P(x) \\
    &= \frac{\varepsilon \Lambda}{5}. \qedhere
  \end{align*}
\end{pf}

Let $F$ be an arbitrary set of $k$ centers.
Partition $\mcal L[i]$ into
\begin{align*}
  \mcal L_{\near}[i] &:= \set*{Z\in \mcal L[i]: \min_{x\in Z} \kappa(x, F)^p \leq \frac5{\varepsilon}(2^{-i+1} \Delta)^p} \\
  \mcal L_{\far}[i] &:= \set*{Z\in \mcal L[i]: \min_{x\in Z} \kappa(x, F)^p > \frac5{\varepsilon}(2^{-i+1} \Delta)^p}.
\end{align*}

\begin{prop}
  The output $\mcal L_{\far}$ of \Cref{alg:reproducible coreset} satisfies
  \[
    \sum_{i=0}^t \int_{\mcal L_{\far}[i]} \kappa(x, R(x))^p d\P(x)
    \leq \frac{\varepsilon}5 \cost(F).
  \]
\end{prop}

\begin{pf}
  Any $x\in Z\in \mcal L_{\far}[i]$ contributes a cost of at least $\frac5{\varepsilon}(2^{-i+1} \Delta)^p$.
  Hence
  \begin{align*}
    \sum_{i=0}^t \int_{\mcal L_{\far}[i]} \kappa(x, R(x))^p d\P(x)
    &\leq \sum_{i=0}^t \int_{\mcal L_{\far}[i]} (2^{-i+1} \Delta)^p d\P(x) \\
    &= \frac{\varepsilon}5 \sum_{i=0}^t \int_{\mcal L_{\far}[i]} \frac5{\varepsilon} (2^{-i+1} \Delta)^p d\P(x) \\
    &\leq \frac{\varepsilon}5 \sum_{i=0}^t \int_{\mcal L_{\far}[i]} \kappa(x, F)^p d\P(x) \\
    &\leq \frac{\varepsilon}5 \cost(F). \qedhere
  \end{align*}
\end{pf}

\begin{prop}
  The output $\mcal L_{\near}$ of \Cref{alg:reproducible coreset} satisfies
  $\card{\mcal L_{\near}[i]} \leq kM$ where
  \[
    M := \left[ \frac{2^5}{\varepsilon} \Delta \right]^d
  \]
\end{prop}

\begin{pf}
  The furthest point $x$ in a cell in $\mcal L_{\near}[i]$
  can have a distance of at most
  \[
    2^{-i} \Delta + \sqrt[p]{\frac5{\varepsilon}}(2^{-i+1} \Delta)
    \leq \left( 1 + \frac{10}{\varepsilon} \right)(2^{-i} \Delta)
  \]
  to the nearest center.
  Thus the points belonging to a particular center
  live in an $\ell_\infty$ ball of that radius.
  Not including the cell containing the center,
  we can walk past at most
  \[
    \frac1{2^{-i}}\cdot \left( 1 + \frac{10}{\varepsilon} \right)(2^{-i} \Delta)
    = \left( 1 + \frac{10}{\varepsilon} \right) \Delta
  \]
  cells if we walk in the directions of the canonical basis of $\R^d$.
  It follows that there can be at most
  \[
    \left[ 1 + 2\left( 1 + \frac{10}{\varepsilon} \right) \Delta \right]^d
    \leq \left[ \frac{32}{\varepsilon} \Delta \right]^d
  \]
  cells in $\mcal L_{\near}[i]$ close to each center of $F$,
  and thus there are at most $kM$ cells in total.
\end{pf}

\begin{prop}
  By choosing
  \[
    \gamma := \frac{\varepsilon}{5t kM (2\Delta)^p},
  \]
  The output $\mcal L_{\near}$ of \Cref{alg:reproducible coreset} satisfies
  \[
    \sum_{i=0}^t \int_{\mcal L_{\near}[i]} \kappa(x, R(x))^p
    \leq \frac{\varepsilon \Lambda}{5}.
  \]
\end{prop}

\begin{pf}
  Each light cell has measure at most $\frac{\gamma\cdot \Lambda}{2^{-pi}}$.
  By computation,
  \begin{align*}
    \sum_{i=0}^t \int_{\mcal L_{\near}[i]} \kappa(x, R(x))^p d\P(x)
    &\leq \sum_{i=1}^t \int_{\mcal L_{\near}[i]} (2^{-i+1} \Delta)^p d\P(x) \\
    &\leq \sum_{i=1}^t kM\cdot \frac{\gamma\cdot \Lambda}{2^{-pi}} 2^{-pi} (2\Delta)^p \\
    &= t\cdot kM\cdot \gamma\cdot \Lambda \cdot (2\Delta)^p \\
    &\leq \frac{\varepsilon \Lambda}{5}. \qedhere
  \end{align*}
\end{pf}

\begin{cor}\label{cor:expected shift}
  Set
  \[
    \gamma := \frac{\varepsilon}{5t kM (2\Delta)^p}.
  \]
  Then for any set $F$ of $k$-centers,
  the output $R$ of \Cref{alg:reproducible coreset} satisfies
  \[
    \int_{\mcal X} \kappa(x, R(x))^p d\P(x)
    \leq \varepsilon \cost(F).
  \]
\end{cor}

\begin{pf}
  $\mcal X$ is contained in the union of all $\mcal L[i]$'s and $\mcal S[t]$.
  Hence
  \begin{align*}
    \int_{\mcal X} \kappa(x, R(x))^p d\P(x)
    &\leq \frac{2\varepsilon \Lambda}{5} + \frac{\varepsilon}5 \cost(F) \\
    &\leq \frac{2\varepsilon (1+\varepsilon) \OPT}{5} + \frac{\varepsilon}5 \cost(F) \\
    &\leq \frac{4\varepsilon}5 \OPT + \frac{\varepsilon}5 \cost(F) \\
    &\leq \varepsilon \cost(F). \qedhere
  \end{align*}
\end{pf}

\subsubsection{Coreset Size}\label{sec:coreset size}
To determine the size of our coreset,
we bound the number of marked heavy cells,
as all our representative points are in marked heavy cells.

Fix $F_{\OPT}$ to be a set of optimal centers for the $(k, p)$-clustering problem.
Write $\mcal M[i]$ to denote the set of marked heavy cells on the $i$-th layer
and partition $\mcal M[i]$ as
\begin{align*}
  \mcal M_{\close}[i] &:= \set*{Z\in \mcal M[i]: \min_{x\in Z} \kappa(x, F) \leq 2^{-pi+1}} \\
  \mcal M_{\dist}[i] &:= \set*{Z\in \mcal M[i]: \min_{x\in Z} \kappa(x, F) > 2^{-pi+1}}.
\end{align*}

\begin{prop}
  The marked heavy cells $\mcal M_{\dist}$ outputted by \Cref{alg:reproducible coreset} satisfy
  \[
    \card*{\bigcup_{0\leq i\leq t} \mcal M_{\dist}[i]} \leq \frac{2\beta}\gamma.
  \]
\end{prop}

\begin{pf}
  Each cell in $\mcal M_{\dist}[i]$ has mass at least $\frac{\gamma\cdot \Lambda}{2^{-pi+1}}$
  and each $x\in Z\in \mcal M_{\dist}[i]$ is of distance at least $2^{-pi+1}$ to its closest center.
  Thus each cell contributes at least
  \[
    \frac{\gamma\cdot \Lambda}{2^{-pi+1}}\cdot 2^{-pi+1} = \gamma \cdot \Lambda
  \]
  to the objective.
  If follows that there are at most
  \[
    \frac{\OPT}{\gamma \Lambda}
    \leq \frac1\gamma \beta \left( 1+\varepsilon \right)
    \leq \frac{2\beta}\gamma.
  \]
  such cells.
\end{pf}

\begin{prop}
  The marked heavy cells $\mcal M_{\close}$ outputted by \Cref{alg:reproducible coreset} satisfy
  \[
    \card*{\mcal M_{\close}[i]} \leq k (7\Delta)^d.
  \]
\end{prop}

\begin{pf}
  The furthest point $x$ in a cell $\mcal M_{\close}[i]$ to its nearest center is
  \[
    2^{-pi+1} + 2^{-i} \Delta.
  \]
  In other words,
  not including the cell containing the center,
  we can walk past at most
  \[
    2^{i(1-p)+1} + \Delta
    \leq 2 + \Delta
  \]
  cells by walking along the direction of the canonical basis of $\R^d$.
  Thus there are at most
  \[
    k\left[ 1 + 2\left( 2 + \Delta \right) \right]^d
    \leq k\left[ 7\Delta \right]^d
  \]
  such cells on the $i$-th layer.
\end{pf}

\begin{prop}\label{prop:coreset size}
  The image of the map $R$ outputted by \Cref{alg:reproducible coreset} satisfies
  \[
    N := \card{R(\mcal X)}
    \leq \frac{2\beta}\gamma + tk (7\Delta)^d.
  \]
  In particular,
  for 
  \[
    \gamma := \frac{\varepsilon}{5t kM (2\Delta)^p},
  \]
  we have
  \[
    N := \card{R(\mcal X)}
    \leq \frac{3\beta}\gamma
    = 3\beta\cdot\frac{5t kM (2\Delta)^p}\varepsilon.
  \]
\end{prop}

\begin{pf}
  The image is contained in the marked heavy cells.
\end{pf}

\subsubsection{\texorpdfstring{$k$}{k}-Medians \& \texorpdfstring{$k$}{k}-Means}\label{sec:coreset summary}
We now derive results for the case of $p=1$.
\begin{thm}\label{thm:k-medians coreset}
  \Cref{alg:reproducible coreset} is $\rho$-replicable
  and outputs an $\varepsilon$-coreset
  of size
  \[
    O\left( \frac{k 2^{5d} \Delta^{d+1}}{\varepsilon^{d+1}} \log \frac{\Delta}{\varepsilon\cdot \OPT} \right)
  \]
  for statistical $k$-medians
  with probability at least $1-\delta$.
  Moreover,
  it has sample complexity
  \[
    \tilde O\left( \frac{k^3 2^{15d} \Delta^{3d+3}}{\rho^2 \varepsilon^{3d+3}\cdot \OPT^3}\log\frac1\delta \right).
  \]
\end{thm}

\begin{pf}[\Cref{thm:k-medians coreset}]
  By \Cref{cor:expected shift},
  \Cref{alg:reproducible coreset} outputs an $\varepsilon$-coreset.
  By \Cref{cor:coreset sample complexity},
  \Cref{alg:reproducible coreset} is $\rho$-replicable with sample complexity
  \[
    \tilde O \left( \frac{t^2}{\rho^2\gamma^3\cdot \OPT^3}\log\frac1\delta \right).
  \]

  We have
  \[
    \frac{3\beta}{\gamma}
    = O\left( \frac{k \left[ \nicefrac{2^5 \Delta}\varepsilon  \right]^d (2\Delta)}{\varepsilon}\cdot \log \frac{\Delta}{\varepsilon\cdot \OPT} \right)
    = O\left( \frac{k 2^{5d} \Delta^{d+1}}{\varepsilon^{d+1}} \log \frac{\Delta}{\varepsilon\cdot \OPT} \right)
  \]
  By \Cref{prop:coreset size},
  the coreset has size at most this value.

  Substituting the value of $\gamma$ above
  and remarking that 
  $t$ is a polylogarithmic factor of the other terms,
  we thus conclude that the final sample complexity is
  \[
    \tilde O \left( \frac{k^3 2^{15d} \Delta^{3d+3}}{\varepsilon^{3d+3}}\cdot \frac{1}{\rho^2\cdot \OPT^3}\log\frac1\delta \right)
    = \tilde O\left( \frac{k^3 2^{15d} \Delta^{3d+3}}{\rho^2 \varepsilon^{3d+3}\cdot \OPT^3}\frac1\delta \right). \qedhere
  \]
\end{pf}

Now consider the case of $p=2$.
We can perform our analysis with the help of a standard inequality.

\begin{prop}\label{prop:k-means inequality}
  For any set of centers $F$,
  \begin{align*}
    \E_x \kappa(R(x), F)^2
    &\leq \E_x \kappa(x, F)^2
    + 2\sqrt{\E_x\left[ \kappa(x, F)^2 \right]\cdot \E_x\left[ \kappa(x, R(x))^2 \right]}
    + \E_x \kappa(x, R(x))^2 \\
    \E_x \kappa(x, F)^2
    &\leq \E_x \kappa(R(x), F)^2
    + 2\sqrt{\E_x\left[ \kappa(R(x), F)^2 \right]\cdot \E_x\left[ \kappa(x, R(x))^2 \right]}
    + \E_x \kappa(x, R(x))^2.
  \end{align*}

  Thus if $\E_x \kappa(x, R(x))^2 \leq \frac{\varepsilon^2}{2^6} \E_x \kappa(x, F)^2$,
  \begin{align*}
    \E_x \kappa(R(x), F)^2
    &\leq \E_x \kappa(x, F)^2
    + 2\cdot \frac1{2^3} \varepsilon\E_x\left[ \kappa(x, F)^2 \right]
    + \frac1{2^6}\varepsilon^2 \E_x\left[ \kappa(x, F)^2 \right] \\
    &\leq \left( 1+\frac\varepsilon4 \right) \E_x \kappa(x, F)^2 \\
    \E_x \kappa(x, F)^2
    &\leq \E_x \kappa(R(x), F)^2
    + 2\sqrt{\left( 1+\varepsilon \right) \E_x\left[ \kappa(x, F)^2 \right]\cdot \frac1{2^6}\varepsilon^2 \E_x\left[ \kappa(x, F)^2 \right]}
    + \frac1{2^6} \varepsilon^2 \E_x \kappa(x, F)^2 \\
    &\leq \E_x \kappa(R(x), F)^2
    + \frac12\varepsilon \E_x\left[ \kappa(x, F)^2 \right]
    + \frac1{2^6}\varepsilon^2 \E_x \kappa(x, F)^2 \\
    &\leq \E_x \kappa(R(x), F)^2 + \varepsilon \E_x \kappa(x, F)^2.
  \end{align*}
\end{prop}

\begin{pf}
  By H\"older's inequality,
  \begin{align*}
    \E_x \kappa(R(x), F)^2
    &\leq \E_x \left[ \kappa(x, F) + \kappa(x, R(x)) \right]^2 \\
    &= \E_x \kappa(x, F)^2 + 2 \E_x [\kappa(x, F)\kappa(x, R(x))] + \E_x \kappa(x, R(x))^2 \\
    &\leq \E_x \kappa(x, F)^2 + 2\sqrt{\E_x \left[ \kappa(x, F)^2 \right] \E_x \left[ \kappa(x, R(x))^2 \right]} + \E_x \kappa(x, R(x))^2.
  \end{align*}

  The other case is identical.
\end{pf}

\begin{thm}\label{thm:k-means coreset}
  \Cref{alg:reproducible coreset} is $\rho$-replicable
  and outputs an $\varepsilon$-coreset
  for statistical $k$-means
  of size
  \[
    O\left( \frac{k 2^{11d} \Delta^{d+2}}{\varepsilon^{2d+2}} \log\frac{\Delta^2}{\varepsilon \OPT} \right)
  \]
  with probability at least $1-\delta$.
  Moreover,
  it has sample complexity
  \[
    \tilde O\left( \frac{k^3 2^{33d} \Delta^{3d+6}}{\rho^2 \varepsilon^{6d+6}\cdot \OPT^3}\log\frac1\delta \right).
  \]
\end{thm}

\begin{pf}[\Cref{thm:k-means coreset}]
  Similar to $k$-medians,
  \Cref{alg:reproducible coreset} outputs an $\varepsilon$-coreset
  with probability at least $1-\delta$,
  is $\rho$-replicable,
  and has sample complexity
  \[
    \tilde O \left( \frac{t^2}{\rho^2\gamma^3\cdot \OPT^3}\log\frac1\delta \right).
  \]
  However,
  we need to take $\varepsilon' := \nicefrac{\varepsilon^2}{2^6}$.

  We have
  \[
    \frac3{\gamma}
    = O\left( \frac{t k \left[ \nicefrac{2^{5+6} \Delta}{\varepsilon^2} \right]^d (2\Delta)^2}{\varepsilon^2} \right)
    = O\left( \frac{k 2^{11d} \Delta^{d+2}}{\varepsilon^{2d+2}} \log\frac{\Delta^2}{\varepsilon \OPT} \right).
  \]
  By \Cref{prop:coreset size},
  the coreset has size at most this value.

  $t$ is a polylogarithmic factor of the rest of the parameters,
  thus we thus conclude that the final sample complexity is
  \[
    \tilde O \left( \frac{k^3 2^{33d} \Delta^{3d+6}}{\varepsilon^{6d+6}}\cdot \frac{1}{\rho^2\cdot \OPT^3} \right)
    = \tilde O\left( \frac{k^3 2^{33d} \Delta^{3d+6}}{\rho^2 \varepsilon^{6d+6}\cdot \OPT^3}\log\frac1\delta \right). \qedhere
  \]
\end{pf}

It is not clear how to compare our guarantees with that of \citet{frahling2005coresets}
as we attempt to minimize the number of samples required from a distribution
assuming only sample access.
On the other hand,
\citet{frahling2005coresets} attempted to minimize the running time of their algorithm
assuming access to the entire distribution.

\subsection{Replicable Multinomial Parameter Estimation}\label{sec:multinomial estimation}
Now that we have reduced the problem to a finite distribution (coreset),
we can think of it as a weighted instance of $(k, p)$-clustering
where the weights are unknown but can be estimated using data.

\begin{prop}\label{prop:mass estimation}
  Enumerate the coreset $R(\mcal X) = r^{(1)}, \dots, r^{(N)}$
  and let $w^{(i)}$ be the probability mass at $r^{(i)}$.
  If we have estimates $\hat w^{(i)}$ satisfying
  \[
    \abs{\hat w^{(i)} - w^{(i)}} < \frac{\varepsilon}{N},
  \]
  then
  \[
    \abs*{\sum_{i=1}^N w^{(i)} \kappa(r^{(i)}, F)^p - \sum_{i=1}^N \hat w^{(i)} \kappa(r^{(i)}, F)^p}
    \leq \varepsilon.
  \]
  for all centers $F$.
\end{prop}

\begin{pf}
  Indeed,
  \begin{align*}
    \abs*{\sum_{i=1}^N w^{(i)} \kappa(r^{(i)}, F)^p - \sum_{i=1}^N \hat w^{(i)} \kappa(r^{(i)}, F)^p}
    &\leq \sum_{i=1}^N \abs{w^{(i)} - \hat w^{(i)}} \kappa(r^{(i)}, F)^p \\
    &\leq \sum_{i=1}^N \abs{w^{(i)} - \hat w^{(i)}}\cdot 1 \\
    &\leq \varepsilon. \qedhere
  \end{align*}
\end{pf}

Let us formulate this as a replicable multinomial parameter estimation problem.
Consider a multinomial distribution $Z = (Z^{(1)}, \dots, Z^{(N)})$ of dimension $N$
with parameters $p^{(1)}, \dots, p^{(N)}$ and $n$.
Note that this can be cast as a more general statistical query problem which we illustrate below.
We allow simultaneous estimation of more general statistical queries $g_1, \dots, g_N: (\mcal X^n, \P)\to \R$ compared to \citet{impagliazzo2022reproducibility},
assuming they can be estimated with high accuracy and confidence from data,
say each query $g_j(x_1,  \dots, x_n)$ concentrates about its true value $G_j\in \R$ with high probability.

\begin{algorithm}[h]
\caption{Replicable Rounding}\label{alg:reproducible rounding}
\begin{algorithmic}[1]
  \STATE {\bfseries rRounding}(statistical queries $g_1, \dots g_N$, distribution $\P$, number of samples $n$, error $\varepsilon$, replicability $\rho$, confidence $\delta$):
    \STATE Sample $x_1, \dots, x_n \sim \P$
    \STATE $\alpha \gets \frac{2\varepsilon}{\rho + 1 - 2\delta}$
    \FOR {$j\gets 1, \dots, N$:}
      \STATE $\alpha^{(j)} \gets \text{uniform random sample $[0, \alpha]$}$
      \STATE Split $\R$ into intervals $I^{(j)} = \set{[\alpha^{(j)} + z \alpha, \alpha^{(j)} + (z+1) \alpha): z\in \Z}$
      \STATE Output the midpoint $\hat G_j$ of the interval within $I^{(i)}$ that $g_j(x_1, \dots, x_n)$ falls into
    \ENDFOR
\end{algorithmic}
\end{algorithm}

\begin{thm}[Replicable Rounding]\label{thm:reproducible rounding}
  Suppose we have a finite class of statistical queries $g_1, \dots, g_N$
  and sampling $n$ independent points from $\P$ ensures that
  \[
    \sum_{j=1}^N \abs{g_j(x_1, \dots, x_n) - G_j} \leq \varepsilon' := \frac{\varepsilon(\rho - 2\delta)}{\rho+1-2\delta}
  \]
  with probability at least $1-\delta$.

  Then \Cref{alg:reproducible rounding} is $\rho$-replicable
  and outputs estimates $\hat G_j$
  such that
  \[
    \abs{\hat G_j - G_j}\leq \varepsilon
  \]
  with probability at least $1-\delta$ for every $j\in [N]$.
  Moreover,
  it requires at most $n$ samples.
\end{thm}

\begin{pf}
  Outputting the midpoint of the interval can offset each $g_j(x_1, \dots, x_n)$
  by at most $\frac\alpha2$.
  Hence
  \[
    \abs{\hat G_j - G_j} \leq \frac{\varepsilon(\rho - 2\delta)}{\rho + 1 - 2\delta} + \frac{\varepsilon}{\rho+1-2\delta} = \varepsilon.
  \]

  Consider two executions of our algorithm
  yielding estimates $\hat G^{(1)}, \hat G^{(2)}$.
  The probability that any of the estimates fail to satisfy $\ell_1$-tolerance $\varepsilon'$
  is at most $2\delta$.
  The two executions output different estimates
  only if some random offset ``splits'' some $\hat G_j^{(1)}, \hat G_j^{(2)}$.
  Conditioning on the success to satisfy $\ell_1$-tolerance,
  this occurs with probability at most
  \begin{align*}
    \sum_{j=1}^N \frac{\abs{\hat G_j^{(1)} - \hat G_j^{(2)}}}{\alpha}
    &\leq \sum_{j=1}^N \frac{\abs{\hat G_j^{(1)} - G_j} + \abs{\hat G_j^{(2)} - G_j}}{\alpha} \\
    &\leq \frac{2\varepsilon'}{\alpha} \\
    &\leq \rho - 2\delta.
  \end{align*}
  Accounting for the probability of failure to satisfy the $\ell_1$-tolerance of $2\delta$,
  our algorithm is $\rho$-replicable.
\end{pf}

We note that \Cref{thm:reproducible rounding}
can be thought of as a generalization of the SQ oracle (cf. \Cref{thm:reproducible sq oracle}) from \citet{impagliazzo2022reproducibility}.
Indeed,
simply using \Cref{thm:reproducible sq oracle} yields an extra factor of $N$ in the sample complexity
as we must take a union bound.

\begin{cor}\label{cor:replicable mass estimation}
  Let $\varepsilon, \rho\in (0, 1)$ and $\delta\in (0, \nicefrac\rho3)$.
  There is a $\rho$-replicable algorithm
  that outputs parameter estimates $\bar p$ for a multinomial distribution of dimension $N$
  such that
  \begin{enumerate}[(a)]
        \item $\abs{\bar p^{(i)} - p^{(i)}}\leq \varepsilon$
        for every $i\in [N]$
        with probability at least $1-\delta$.
        \item $\bar p^{(i)}\geq 0$ for all $i\in [N]$.
        \item $\sum_i \bar p^{(i)} = 1$.
  \end{enumerate}
  Moreover,
  the algorithm has sample complexity
  \[
    O\left( \frac{\ln\nicefrac1\delta + N}{\varepsilon^2 (\rho-\delta)^2} \right)
    = \tilde O\left( \frac{N}{\varepsilon^2 \rho^2}\log\frac1\delta \right).
  \]
\end{cor}

\begin{pf}
  Define
  \[
    \varepsilon' := \frac{\varepsilon(\rho - 2\delta)}{\rho + 1 - 2\delta}.
  \]

  By \Cref{prop:BHC inequality},
  sampling
  \begin{align*}
    n &= \frac{2\ln\nicefrac1\delta + 2N\ln 2}{\varepsilon'^2} \\
    &\leq \frac{8\ln\nicefrac1\delta + 8N\ln 2}{\varepsilon^2 (\rho-2\delta)^2}.
  \end{align*}
  points implies that $\sum_{i=1}^N \abs{\wh p^{(i)} - p^{(i)}} < \varepsilon'$
  with probability at least $1-\delta$.
  
  Thus running \Cref{alg:reproducible rounding}
  with functions
  \[
    g_j(x_1, \dots, x_N) := \frac1N \sum_{i=1}^N \ones\set{x_i=j}
  \]
  yields $\bar p^{(i)}$'s such that $\abs{\bar p^{(i)} - p^{(i)}} \leq \varepsilon$
  for each $i\in [N]$.
  If there are any negative $\bar p^{(i)}$'s,
  we can only improve the approximation by setting them to 0.
  If the sum of estimates is not equal to 1,
  we can normalize by taking
  \[
    \bar p^{(i)} \gets \bar p^{(i)} - \frac1N\left[ \sum_i \bar p^{(i)} - 1 \right].
  \]
  This introduces an additional error of $\nicefrac\varepsilon{N}$ for each $\hat p^{(i)}$
  and a maximum error of $2\varepsilon$.
  Choosing $\varepsilon_1 := \nicefrac\varepsilon2$ concludes the proof.
\end{pf}

\subsection{Replicable OPT Estimation for \texorpdfstring{$(k, p)$}{(k, p)}-Clustering}\label{sec:k-medians OPT estimation}
Our coreset algorithm assumes the knowledge of a constant ratio estimation of OPT.
If this is not provided to us,
we show in this section that it is possible to replicably compute such an estimate
using a two-step approach.
First,
we are able to produce replicable estimates with additive error $\varepsilon$ simply by approximately solving the sample $(k, p)$-clustering problem on a sufficiently large sample thanks to uniform convergence (cf. \Cref{thm:uniform convergence}).
Then,
we repeat this process with $\varepsilon \gets \nicefrac\varepsilon2$ until $\varepsilon$ is but a small fraction of the outputted estimate.

\begin{thm}[OPT Estimation with Additive Error]
    Let $\beta \geq 1$ be an absolute constant
    and $\varepsilon, \rho\in (0, 1), \delta\in (0, \nicefrac\rho3)$.
    Suppose we are provided with an algorithm $\mcal A$ that,
    given an instance of sample $(k, p)$-clustering,
    outputs an estimate $\Xi$ of the value $\wh \OPT$
    such that
    \[
        \Xi\in \left[ \frac1\beta \wh \OPT, \wh\OPT \right].
    \]
    Then there is a $\rho$-replicable algorithm
    which produces an estimate $\Lambda$ of $\OPT$
    such that
    \[
        \Lambda\in \left[ \frac{\OPT}\beta - \varepsilon, \OPT + \varepsilon \right]
    \]
    with probability at least $1-\delta$.
    Moreover,
    the algorithm has sample complexity
    \[
      \tilde O\left( \frac{k^2d^2}{\varepsilon^6\rho^6}\log\frac{p}\delta \right).
    \]
\end{thm}

We note that the algorithm $\mcal A$ can be taken to be any $\beta$-approximation algorithm
with some postprocessing,
where we divide the value of the cost by $\beta$.
Alternatively,
we can take $\mcal A$ to be some convex relaxation of our problem with constant integral gap.

\begin{pf}
    Let $X_1, \dots, X_N\sim \P$ be i.i.d. random variables.
    The output $\Xi = \mcal A(X_1, \dots, X_N)\in [0, 1]$ is thus a bounded random variable.
    Suppose we repeat this experiment $n$ times to produce estimates $\Xi_1, \dots, \Xi_n$.
    By an Hoeffding bound,
    \[
        \P\set*{\frac1n \sum_{i=1}^n (\Xi_i - \E[\Xi]) > \varepsilon}
        \leq 2\exp\left( -2n\varepsilon^2 \right).
    \]
    Thus with
    \[
        n\geq \frac1{2\varepsilon^2} \ln\frac2\delta
    \]
    trials,
    the average estimate $\bar \Xi := \frac1n \sum_{i=1}^n \Xi_i$ satisfies $\abs*{\bar \Xi - \E[\Xi]} < \varepsilon$
    with probability at least $1-\delta$.

    Now,
    by \Cref{thm:uniform convergence},
    choosing
    \[
        N = O\left( \frac{k^2 d^2}{\varepsilon^4} \log\frac{np}\delta \right)
    \]
    i.i.d. samples from $\P$
    ensures that $\abs*{\wh \OPT - \OPT} \leq \varepsilon$
    for all $n$ trial
    with probability at least $1-\delta$.
    Thus conditioning on success,
    we have
    \[
        \Xi_i\in \left[ \frac{\OPT}\beta - \varepsilon, \OPT+\varepsilon \right]
    \]
    for every trial $i\in [n]$.
    But then the average $\bar \Xi$ also falls into this interval as well.

    We have shown that there is an algorithm that outputs an estimate
    \[
        \bar\Xi \in [\E[\Xi] - \varepsilon, \E[\Xi] + \varepsilon] \cap \left[ \frac{\OPT}\beta - \varepsilon, \OPT+\varepsilon \right]
    \]
    with probability at least $1-\delta$.
    Moreover,
    it has sample complexity
    \begin{align*}
        nN &= O\left( \frac1{\varepsilon^2} \log\frac1\delta\cdot \frac{k^2d^2}{\varepsilon^4}\log \left( \frac{p}{\varepsilon^2 \delta}\log\frac1\delta \right) \right) \\
        &= O\left( \frac{k^2d^2}{\varepsilon^6}\log^2 \frac{p}{\varepsilon \delta} \right)
    \end{align*}

    We can now apply the replicable rounding algorithm (cf. \Cref{alg:reproducible rounding}) to achieve the desired outcome.
    Indeed,
    by \Cref{thm:reproducible rounding},
    the output $\Lambda$ after the rounding procedure is $\rho$-reproducible
    and offsets the average $\bar\Xi$ by at most $\varepsilon$.
    Hence we have
    \[
        \Lambda \in \left[ \frac{\OPT}\beta - 2\varepsilon, \OPT+2\varepsilon. \right]
    \]
    with probability at least $1-\delta$.
    Finally,
    the algorithm has sample complexity
    \[
        O\left( \frac{k^2d^2}{\varepsilon^6\rho^6}\log^2 \frac{p}{\varepsilon \rho \delta} \right)
    \]

    Choosing $\varepsilon' = \nicefrac\varepsilon2$ completes the proof.
\end{pf}

\begin{thm}[OPT Estimation with Relative Error]\label{thm:opt estimation relative error}
  Fix $\varepsilon, \rho\in \left( 0, 1 \right)$ and $\delta\in (0, \nicefrac\rho3)$.
  There is a $\rho$-replicable algorithm
  such that with probability at least $1-\delta$,
  it outputs an estimate $\Lambda$ of $\OPT$
  where
  \begin{align*}
    \Lambda &\in \left[ \frac1{\beta (1+\varepsilon)} \OPT, (1+\varepsilon) \OPT \right].
  \end{align*}
  Moreover,
  it has sample complexity
  \[
    \tilde O\left( \frac{k^2d^2 \beta^{12}}{\varepsilon^{12}\rho^6\cdot \OPT^{12}}\log\frac{p}{\delta} \right).
  \]
\end{thm}

\begin{pf}
  We run the estimation algorithm with additive error with
  \begin{align*}
    \varepsilon_i &:= 2^{-i} \\
    \rho_i &:= 2^{-i} \rho \\
    \delta_i &:= 2^{-i} \delta
  \end{align*}
  for $i=1, 2, 3, \dots$ until we obtain an estimate $\Lambda_i$ such that $\varepsilon_i\leq \nicefrac{\varepsilon \Lambda_i}2$.

  Remark that
  \begin{align*}
    &2^{-i} \leq \frac12 \varepsilon \Lambda_i \\
    &\impliedby 2^{-i} \leq \frac12 \varepsilon \left( \frac\OPT\beta - 2^{-i} \right) \\
    &\iff 2^{-i}\left( 1 + \frac12 \varepsilon \right) \leq \frac{\varepsilon \OPT}{2\beta} \\
    &\impliedby -i \leq \log \frac{\varepsilon \OPT}{4\beta} \\
    &\iff i = \log\frac{\beta}{\varepsilon \OPT} + 2.
  \end{align*}
  Thus the algorithm certainly terminates.

  Upon termination,
  we output some $\Lambda$ such that $\Lambda\in \left[ \nicefrac{\OPT}\beta - \nicefrac{\varepsilon \Lambda}2, \OPT + \nicefrac{\varepsilon \Lambda}2 \right]$.
  It follows that
  \begin{align*}
    \Lambda &\leq \OPT + \frac12\varepsilon \Lambda \\
    \left( 1-\frac12\varepsilon \right) \Lambda &\leq \OPT \\
    \Lambda &\leq \frac1{1-\frac\varepsilon2} \OPT \\
    &\leq (1+\varepsilon) \OPT
  \end{align*}
  and
  \begin{align*}
    \Lambda &\geq \frac{\OPT}\beta - \frac12\varepsilon \Lambda \\
    \beta \left( 1 + \frac12\varepsilon \right) \Lambda &\geq \OPT \\
    \Lambda &\geq \frac1{\beta(1+\varepsilon)} \OPT.
  \end{align*}

  The probability of not being replicable in each iteration is $2^{-i} \rho$.
  Hence the total probability is at most
  \[
    \sum_{i=1}^\infty 2^{-i} \rho = \rho
  \]
  and similarly for $\delta$.

  Finally,
  there are $O\left( \log\frac\beta{\varepsilon \OPT} \right)$ iterations
  and the sample complexity of each iteration is bounded above
  by the sample complexity of the final iteration
  where
  \[
    2^{-i} = \Omega\left( \frac{\varepsilon\OPT}\beta \right).
  \]
  Hence the total sample complexity is
  \[
    \tilde O\left( \log\frac\beta{\varepsilon \OPT}\cdot \frac{k^2d^2}{\left( \frac{\varepsilon\OPT}\beta \right)^6 \left( \frac{\varepsilon\OPT}\beta\cdot \rho\right)^6}\log\frac{p}\delta \right)
    =  \tilde O\left( \frac{k^2d^2 \beta^{12}}{\varepsilon^{12}\rho^6\cdot \OPT^{12}}\log\frac{p}\delta \right). \qedhere
  \]
\end{pf}

\subsection{Putting it Together}\label{sec:statistical clustering summary}
We are now ready to prove the results on statistical $k$-medians
and statistical $k$-means,
which we restate below for convenience.
Recall that $\beta$ is an absolute constant
and hence disappears under the big-O notation.

\kMedians*

\begin{pf}[\Cref{thm:statistical k-medians algorithm}]
  First,
  we compute $\Lambda$,
  an estimate of $\OPT$ with relative error $\varepsilon$,
  confidence $\nicefrac\delta3$,
  and replicability parameter $\nicefrac{\rho}3$.
  Then,
  we produce an $\nicefrac{\varepsilon}2$-coreset for $k$-medians of cardinality $N$
  with confidence and replicability parameters $\nicefrac\delta3, \nicefrac\rho3$.
  Finally,
  we estimate the mass at each point of the coreset
  with confidence and replicability parameters $\nicefrac\delta3, \nicefrac\rho3$
  to an accuracy of
  \[
    \frac{\varepsilon \Lambda}{4N}
    \leq \frac{\varepsilon \OPT}{2N}.
  \]
  By \Cref{prop:mass estimation},
  running the approximation oracle on the coreset with the estimated weights
  yields a $\beta(1+\varepsilon)$-approximation.

  By \Cref{thm:opt estimation relative error},
  computing $\Lambda$ requires
  \[
    \tilde O\left( \frac{k^2d^2}{\varepsilon^{12}\rho^6\cdot \OPT^{12}}\log\frac1\delta \right)
  \]
  samples.

  By \Cref{thm:k-medians coreset},
  the coreset construction yields a coreset of cardinality
  \[
    N = O\left( \frac{k 2^{5d} \Delta^{d+1}}{\varepsilon^{d+1}} \log \frac{\Delta}{\varepsilon\cdot \OPT}\cdot 2^d \right)
    = O\left( \frac{k 2^{6d} \Delta^{d+1}}{\varepsilon^{d+1}} \log \frac{\Delta}{\varepsilon\cdot \OPT} \right)
  \]
  and incurs a sample cost of
  \[
    \tilde O\left( \frac{k^3 2^{15d} \Delta^{3d+3}}{\rho^2 \varepsilon^{3d+3}\cdot \OPT^3}\cdot 2^{3d}\frac1\delta \right)
    = \tilde O\left( \frac{k^3 2^{18d} \Delta^{3d+3}}{\rho^2 \varepsilon^{3d+3}\cdot \OPT^3}\frac1\delta \right).
  \]
  
  Finally,
  \Cref{cor:replicable mass estimation} states that the probability mass estimation incurs a sample cost of
  \begin{align*}
    \tilde O\left( \frac{N^3}{\varepsilon^2 \rho^2\cdot \Lambda^2}\log\frac1\delta \right)
    &= \tilde O\left( \frac{N^3}{\varepsilon^2 \rho^2\cdot \OPT^2}\log\frac1\delta \right) \\
    &= \tilde O\left( \frac{k^3 2^{18d} \Delta^{3d+3}}{\varepsilon^{3d+3}} \cdot \frac{1}{\varepsilon^2 \rho^2\cdot \OPT^2}\log\frac1\delta \right) \\
    &= \tilde O\left( \frac{k^3 2^{18d} \Delta^{3d+3}}{\rho^2 \varepsilon^{3d+5}\cdot \OPT^2}\log\frac1\delta \right). \qedhere
  \end{align*}
\end{pf}

\kMeans*

\begin{pf}[\Cref{thm:statistical k-means algorithm}]
  Similar to $k$-median,
  we first compute $\Lambda$,
  an estimate of $\OPT$ with relative error $\varepsilon$,
  confidence $\nicefrac\delta3$,
  and replicability parameter $\nicefrac{\rho}3$.
  Then,
  we produce an $\nicefrac{\varepsilon}2$-coreset for $k$-medians of cardinality $N$
  with confidence and replicability parameters $\nicefrac\delta3, \nicefrac\rho3$.
  Finally,
  we estimate the mass at each point of the coreset
  with confidence and replicability parameters $\nicefrac\delta3, \nicefrac\rho3$
  to an accuracy of
  \[
    \frac{\varepsilon \Lambda}{4N}
    \leq \frac{\varepsilon \OPT}{2N}.
  \]
  By \Cref{prop:mass estimation},
  running the approximation oracle on the coreset with the estimated weights
  yields a $\beta(1+\varepsilon)$-approximation.

  By \Cref{thm:opt estimation relative error},
  computing $\Lambda$ requires
  \[
    \tilde O\left( \frac{k^2d^2}{\varepsilon^{12}\rho^6\cdot \OPT^{12}}\log\frac1\delta \right)
  \]
  samples.

  By \Cref{thm:k-means coreset},
  the coreset construction yields a coreset of cardinality
  \[
    N
    = O\left( \frac{k 2^{11d} \Delta^{d+2}}{\varepsilon^{2d+2}} \log\frac{\Delta}{\varepsilon \OPT}\cdot 2^{2d} \right) 
    = O\left( \frac{k 2^{13d} \Delta^{d+2}}{\varepsilon^{2d+2}} \log\frac{\Delta}{\varepsilon \OPT} \right).
  \]
  and incurs a sample cost of
  \[
    \tilde O\left( \frac{k^3 2^{33d} \Delta^{3d+6}}{\rho^2 \varepsilon^{6d+6}\cdot \OPT^3}\cdot 2^{6d} \right)
    = \tilde O\left( \frac{k^3 2^{39d} \Delta^{3d+6}}{\rho^2 \varepsilon^{6d+6}\cdot \OPT^3}\frac1\delta \right).
  \]
  
  Finally,
  \Cref{cor:replicable mass estimation} states that the probability mass estimation incurs a sample cost of
  \begin{align*}
    \tilde O\left( \frac{N^3}{\varepsilon^2 \rho^2\cdot \Lambda^2}\frac1\delta \right)
    &= \tilde O\left( \frac{N^3}{\varepsilon^2 \rho^2\cdot \OPT^2}\frac1\delta \right) \\
    &= \tilde O\left( \frac{k^3 2^{39d} \Delta^{3d+6}}{\varepsilon^{6d+6}} \cdot \frac{1}{\varepsilon^2 \rho^2\cdot \OPT^2}\frac1\delta \right) \\
    &= \tilde O\left( \frac{k^3 2^{39d} \Delta^{3d+6}}{\rho^2 \varepsilon^{6d+8}\cdot \OPT^2}\frac1\delta \right). \qedhere
  \end{align*}
\end{pf}

\section{The Euclidean Metric, Dimensionality Reduction, and \texorpdfstring{$(k, p)$}{(k, p)}-Clustering}\label{apx:dim reduction}
We now build towards a proof for \Cref{thm:euclidean clustering algorithm},
which states the formal guarantees of the replicable $k$-medians and $k$-means algorithms
under the Euclidean distance.
Let us begin by recalling the Johnson-Lindenstrauss lemma.

\begin{thm}[\citep{johnson1984extensions}]\label{thm:jl-lemma}
  There exists a family of random linear maps $\pi_{d, m}$ from $\R^d\to \R^m$
  such that the following hold.
  \begin{enumerate}[(i)]
    \item For every $d\geq 1, \varepsilon, \delta\in (0, \nicefrac12), x\in \R^d$,
      and $m = O\left( \nicefrac1{\varepsilon^2} \log\nicefrac1\delta \right)$,
      \[
        \frac1{1+\varepsilon}\norm{\pi x}_2
        \leq \norm{x}_2
        \leq (1+\varepsilon) \norm{\pi x}_2
      \]
      with probability at least $1-\delta$.
    \item $\pi_{d, m}$ is \emph{sub-Gaussian-tailed},
      thus for every unit vector $x\in \R^d$ and $\varepsilon>0$,
      \[
        \norm{\pi x}_2 \geq 1+\varepsilon
      \]
      with probability at most $\exp(-\Omega(\varepsilon^2 m))$.
  \end{enumerate}
  Furthermore,
  we can take $\pi_{d, m}$ to be the set of random orthogonal projections $\R^d\to \R^m$
  scaled by a factor of $\sqrt{\nicefrac dm}$.
\end{thm}
We say that $\pi\in \pi_{d, m}$ is a \emph{standard random dimension-reduction map}
and write $\P_\pi$ to denote the projected distribution on $\R^m$,
i.e., $\P_\pi(\cdot) = \P(\pi^{-1} (\cdot))$.

As mentioned in \Cref{sec:euclidean metric},
it is not entirely clear how to translate between sets of centers for high-dimensional data
and solutions sets for the compressed data.
For this reason,
our goal is to produce a clustering function $f: \mcal X\to [k]$
which computes the label for any given point in polynomial time.
This also requires us to consider another notion of cost which is translated more easily.

\begin{defn}[Partition Cost]\label{defn:partition cost}
Define the cost of a $k$-partition $\mcal C = (C^{(1)}, \dots, C^{(k)})$ of $\mcal X$ as
\[
  \cost(\mcal C) := \min_{u^{(j)}\in \R^d: j\in [k]} \E_x \left[ \sum_{j=1}^k \ones\set{x\in C^{(j)}}\cdot \kappa(x, u^{(j)})^p \right].
\]
\end{defn}
i.e., we consider the expected cost of picking the optimal centers w.r.t. the underlying
data-generating distribution.
Similarly,
we define the following sample partition cost.

\begin{defn}[Sample Partition Cost]
We define the sample cost of a $k$-partition $\mcal C$ of $x_1, \dots, x_n$ as
\[
  \wh\cost(\mcal C, w) = \min_{u^{(j)}\in \R^d: j\in [k]} \sum_{i=1}^n w_i \sum_{j=1}^k \ones\set{x_i\in C^{(j)}} \kappa(x_i, u^{(j)})^p
\]
where $w\geq 0, \sum_i w_i = 1$.
i.e., we consider the cost that is induced by the distribution specified by $w$ on the samples.
We write $\wh\cost(\mcal C)$ to denote the cost with the uniform distribution on the samples.
\end{defn}

Remark that if $\mcal C$ is induced by $F$,
then $\OPT\leq \cost(\mcal C)\leq \cost(F)$
and similarly for the sample partition cost as well.

\begin{thm}[\citep{makarychev2019performance}]\label{thm:sample dim reduction}
  Let $x_1, \dots, x_n\in \R^d$ be an instance of the Euclidean $(k, p)$-clustering problem.
  Fix $\varepsilon\in (0, \nicefrac14), \delta\in (0, 1)$
  and suppose $\pi: \R^d\to \R^m$ is a standard random dimension-reduction map
  with
  \[
    m = O\left( \frac{p^4}{\varepsilon^2} \log \frac{k}{\varepsilon\delta} \right).
  \]

  Then with probability at least $1-\delta$,
  \begin{align*}
    \frac1{1+\varepsilon} \wh \cost(\mcal C)
    \leq \wh \cost(\pi(\mcal C))
    \leq (1+\varepsilon) \wh \cost(\mcal C)
  \end{align*}
  for every $k$-partition $\mcal C$ of $x_1, \dots, x_n$.
\end{thm}

In essence,
this theorem tells us that if we let the centers of the clusters be the centers of the partition,
then the (sample) cost is preserved in the lower dimensional space.
To the best of our knowledge,
this is the strongest result for dimensionality reduction in (sample) $(k, p)$-clustering
which preserves more than just the costs of optimal centers.

The rest of \Cref{apx:dim reduction} serves to generalize \Cref{thm:sample dim reduction}
for preserving partition costs in the distributional setting.
First,
\Cref{apx:scaling} addresses the issue that a random orthogonal projection can increase the diameter of our data in the worst case
by scaling the original data by an appropriate factor.
Next,
\Cref{apx:preserve partition costs} slightly generalizes \Cref{thm:sample dim reduction}
to the weighted sample clustering setting.
Finally,
\Cref{sec:dimensionality reduction} extends \Cref{thm:sample dim reduction} to the distributional setting,
culminating with \Cref{thm:dim reduction},
which may be of independent interest beyond replicability.

Recall that $\P_\pi$ is the push-forward measure on $\pi(\mcal X)$
induced by $\pi$.
We write $\OPT_\pi$ to denote the cost of an optimal solution to the statistical $(k, p)$-clustering problem
on $(\pi(\mcal X), \P_\pi)$
and $\wh \OPT_\pi$ for the $(k, p)$-clustering problem on $\pi(x_1), \dots, \pi(x_n)$.

\subsection{Scaling}\label{apx:scaling}
We may assume that a standard dimension-reduction map $\pi$ satisfies
$\pi = \sqrt{\nicefrac dm} P$ for some orthogonal projection $P$.
Hence $\pi$ has operator norm at most $\sqrt d$ and
\[
  \pi(\mcal X)\sset \sqrt d\mcal B_m.
\]

In order to apply our existing analysis,
we first scale down the samples in the input space by a factor of $\sqrt{d}$
so that the projected space still lives in a $\kappa$-ball of diameter 1.
This does not affect our approximation guarantees as they are all multiplicative
and hence scale-invariant.
However,
the analysis we perform is on the scaled distribution with optimal cost $\OPT' = d^{-\nicefrac{p}2} \OPT$.
This extra term must be accounted for.

We proceed assuming now that $\mcal X\sset \frac1{\sqrt d} \mcal B_d$
and $\pi(\mcal X)\sset \mcal B_m$.

\subsection{Preserving Partition Costs}\label{apx:preserve partition costs}
Let $\wh \cost(\mcal C, w), \wh \cost(\pi(\mcal C), w)$ be the weighted cost of the 
partition and projected partition, respectively. Recall this means that
the points participate with weights $w$ in the objective. Essentially, the next
result shows that since the projection guarantees do not depend on the number of 
points we are projecting, we can simulate the weights by considering multiple copies
of the points.

\begin{prop}\label{prop:weighted dim reduction}
  Let $x_1, \dots, x_n\in \R^d$ be an instance of the Euclidean $(k, p)$-clustering problem
  and $w_1, \dots, w_n\in \R_+$ be weights which sum to 1.
  Fix $\varepsilon\in (0, \nicefrac14), \delta\in (0, 1)$,
  and suppose $\pi: \R^d\to \R^m$ is a standard random dimension-reduction map
  with
  \[
    m = O\left( \frac{p^4}{\varepsilon^2} \log \frac{k}{\varepsilon\delta} \right) .
  \]

  Then with probability at least $1-\delta$,
  \begin{align*}
    \frac1{1+\varepsilon} \wh \cost(\mcal C, w)
    \leq \wh \cost(\pi(\mcal C), w)
    \leq (1+\varepsilon) \wh \cost(\mcal C, w),
  \end{align*}
  for every $k$-partition $\mcal C$ of $x_1, \dots, x_n$.
\end{prop}

\begin{pf}[\Cref{prop:weighted dim reduction}]
  First suppose that $w_i = \nicefrac{a_i}{b_i}$ for $a_i, b_i\in \Z_+$
  and consider
  \[
    w_i' := \lcm(b_1, \dots, b_n) w_i\in \Z_+
  \]
  obtained from $w_i$
  by multiplying by the least common multiple of the denominators.

  Let $y_1, \dots, y_{n'}$ be the multiset obtained from $x_1, \dots, x_n$
  by taking $w_i'$ multiples of $x_i$.
  The cost of any partition of $x_1, \dots, x_n$ with weights $q_1, \dots, q_n$
  is equal to the cost of the induced partition of $y_1, \dots, y_{n'}$ with uniform weights.
  Thus we can apply \Cref{thm:sample dim reduction} on $y_1, \dots, y_{n'}$ to conclude the proof.

  Now,
  for general $w_i\in \R_+$,
  we remark by the density of the rationals in $\R$
  that there are some $q_1, \dots, q_n\in \Q_+$ which sum to 1
  and satisfy
  \[
    \sum_{i=1}^n \abs{q_i - w_i} < \frac{\varepsilon \min(\wh \OPT, \wh \OPT_\pi)}{2}.
  \]

  Then for any $k$-partition $\mcal C$
  and $u^{(1)}, \dots, u^{(k)}\in \mcal B_d$,
  \begin{align*}
    &\abs*{\sum_{i=1}^n q_i \sum_{j=1}^k \ones\set{x_i\in C^{(j)}} \kappa(x_i, u^{(j)})^p - \sum_{i=1}^n w_i \sum_{j=1}^k \ones\set{x_i\in C^{(j)}} \kappa(x_i, u^{(j)})^p} \\
    &\leq \sum_{i=1}^n \abs{q_i - w_i} \sum_{j=1}^k \ones\set{x_i\in C^{(j)}} \kappa(x_i, u^{(j)})^p \\
    &\leq \sum_{i=1}^n \abs{q_i - w_i} \\
    &\leq \frac{\varepsilon}2 \wh \OPT.
  \end{align*}
  In particular
  \[
    \frac1{1+\varepsilon} \cost(\mcal C, w)
    \leq \cost(\mcal C, q)
    \leq (1+\varepsilon) \cost(\mcal C, w).
  \]

  Similarly,
  for any $v^{(1)}, \dots, v^{(k)}\in \mcal B_m$,
  \begin{align*}
    &\abs*{\sum_{i=1}^n q_i \sum_{j=1}^k \ones\set{x_i\in C^{(j)}} \kappa(\pi(x_i), v^{(j)})^p - \sum_{i=1}^n w_i \sum_{j=1}^k \ones\set{x_i\in C^{(j)}} \kappa(\pi(x_i), v^{(j)})^p} \\
    &\leq \frac{\varepsilon}2 \wh \OPT_\pi,
  \end{align*}
  which implies that
  \[
    \frac1{1+\varepsilon} \cost(\pi(\mcal C), w)
    \leq \cost(\pi(\mcal C), q)
    \leq (1+\varepsilon) \cost(\pi(\mcal C), w).
  \]

  Finally,
  we conclude that
  \begin{align*}
      \cost(\mcal C, w)
      &\leq (1+\varepsilon) \cost(\mcal C, q) \\
      &\leq (1+\varepsilon)^2 \cost(\pi(\mcal C), q) \\
      &\leq (1+\varepsilon)^3 \cost(\pi(\mcal C), w) \\
      \cost(\pi(\mcal C), w)
      &\leq (1+\varepsilon)^3 \cost(\mcal C, w).
  \end{align*}
  Choosing $\varepsilon'$ to be a constant fraction of $\varepsilon$ concludes the proof.
\end{pf}

\begin{cor}\label{cor:sample OPT preservation}
  Let $x_1, \dots, x_n\in \R^d$ be an instance of the weighted Euclidean $(k, p)$-clustering problem
  with non-negative weights $w_1, \dots, w_n\in \R_+$ which sum to 1.
  Fix $\varepsilon\in (0, \nicefrac14), \delta\in (0, 1)$
  and suppose $\pi: \R^d\to \R^m$ is a standard random dimension-reduction map
  with
  \[
    m = O\left( \frac{p^4}{\varepsilon^2} \log \frac{k}{\varepsilon\delta} \right).
  \]

  Then with probability at least $1-\delta$,
  \begin{align*}
    \frac1{1+\varepsilon} \wh \OPT
    \leq \wh \OPT_\pi
    \leq (1+\varepsilon) \wh \OPT.
  \end{align*}
\end{cor}

\begin{pf}[\Cref{cor:sample OPT preservation}]
  The optimal clustering cost coincides with the cost of the partition induced by the optimal centers.
\end{pf}

\subsection{Dimensionality Reduction}\label{sec:dimensionality reduction}
Let $G = \set{g^{(1)}, \dots, g^{(k)}}$ be a $\beta$-approximate solution
to the statistical Euclidean $(k, p)$-clustering problem on $(\pi(\mcal X), \P_\pi)$.
We would like to argue that the partition of $\mcal X$ induced by $G$ is a $(1+\varepsilon)\beta$-approximate partition
with probability at least $1-\delta$.
To do so,
we first approximate $\mcal X$ with a finite set $\tilde{\mcal X}$ in a replicable fashion
while preserving partition costs.
This allows us to apply \Cref{prop:weighted dim reduction}.

Recall we write $\Lambda$ to denote a replicable estimate of $\OPT$
such that $\Lambda \in [ \nicefrac{\OPT}{\beta (1+\varepsilon)}, (1+\varepsilon) \OPT ]$ (cf. \Cref{thm:opt estimation relative error}).
Consider a grid of length $\nicefrac{\varepsilon \Lambda}{(4 p \Delta)}$.
for a point $x\in \mcal X$,
we assign it to the center of the cell that it belongs to,
say $\tilde x$.
Let $\tilde{\mcal X} := \set{\tilde x: x\in \mcal X}$ denote this finite set
and $\tilde \P$ the distribution induced by the discretization.
Remark that $\kappa(x, \tilde x)\leq \nicefrac{\varepsilon \Lambda}{(4 p)}$ for every $x\in \mcal X$.

Note that a partition of $\mcal X$ induces a partition of $\tilde{\mcal X}$
and vice versa.
We proceed without making any distinction between the two.

\begin{prop}
  For any point $u\in \mcal B_d$,
  \[
    \abs{\kappa(x, u)^p - \kappa(\tilde x, u)^p}
    \leq p \abs{\kappa(x, u) - \kappa(\tilde x, u)}.
  \]
\end{prop}

\begin{pf}
  By the mean value theorem,
  the function $g: [0, 1]\to \R$ given by $z\mapsto z^p$
  is $p$-Lipschitz:
  \begin{align*}
    \abs{g(y) - g(z)}
    &\leq \sup_{\xi\in [0, 1]} g'(\xi) \abs{y - z} \\
    &\leq p \abs{y - z}. \qedhere
  \end{align*}
\end{pf}

We write $\widetilde{\cost}(\mcal C)$ to denote the cost of the partition $\mcal C$ of $\tilde{\mcal X}$.

\begin{prop}\label{prop:discretization cost preservation}
  For any partition $\mcal C$ of $\mcal X$
  \[
    \frac1{1+\varepsilon} \cost(\mcal C)
    \leq \widetilde{\cost}(\mcal C)
    \leq (1+\varepsilon) \cost(\mcal C).
  \]
\end{prop}

\begin{pf}
  We have
  \begin{align*}
    &\abs*{\E_x\left[ \sum_{j=1}^k \ones\set{x\in C^{(j)}}\cdot \kappa(x, u^{(j)})^p \right] - \E_x\left[ \sum_{j=1}^k \ones\set{x\in C^{(j)}}\cdot \kappa(\tilde x, u^{(j)})^p \right]} \\
    &\leq \E_x\left[ \sum_{j=1}^k \ones\set{x\in C^{(j)}}\cdot \abs{\kappa(x, u^{(j)})^p - \kappa(\tilde x, u^{(j)})^p} \right] \\
    &\leq \E_x\left[ \sum_{j=1}^k \ones\set{x\in C^{(j)}}\cdot p\abs{\kappa(x, u^{(j)}) - \kappa(\tilde x, u^{(j)})} \right] \\
    &\leq p\cdot \frac{\varepsilon \Lambda}{4p} \\
    &\leq \frac{\varepsilon}2 \OPT. \qedhere
  \end{align*}
\end{pf}

\begin{cor}\label{cor:discretization dim reduction opt preservation}
  Fix $\varepsilon\in (0, \nicefrac14), \delta\in (0, 1)$
  and suppose $\pi: \R^d\to \R^m$ is a standard random dimension-reduction map
  with
  \[
    d = O\left( \frac{p^4}{\varepsilon^2} \log \frac{k}{\varepsilon\delta} \right).
  \]

  We have
  \[
    \frac1{(1+\varepsilon)} \OPT
    \leq \widetilde{\OPT}_\pi
    \leq (1+\varepsilon) \OPT \,,
  \]
  with probability at least $1-\delta,$ where the 
  probability is taken w.r.t. the choice of the
  random map.
\end{cor}

\begin{pf}
  The optimal clustering cost coincides with the cost of the partition induced by the optimal centers.
  An application of \Cref{prop:discretization cost preservation} yields the desired result.
\end{pf}

\begin{thm}\label{thm:dim reduction}
  Fix $\varepsilon\in (0, \nicefrac14), \delta\in (0, 1)$
  and suppose $\pi\in \pi_{d, m}$ is a standard random dimension-reduction map with
  \[
    m = O\left( \frac{p^4}{\varepsilon^2} \log \frac{k}{\varepsilon\delta} \right).
  \]
  Let $G = \set{g^{(1)}, \dots, g^{(k)}}$ be a $\beta$-approximate solution
  to the statistical Euclidean $(k, p)$-clustering problem on $(\pi(\tilde{\mcal X}), \tilde{\P}_\pi)$.
  Then the partition of $\mcal X$ induced by $G$ is a $(1+\varepsilon)\beta$-approximate partition
  with probability at least $1-\delta$.
  
  That is,
  for $\mcal C = \set{C^{(1)}, \dots, C^{(k)}}$ given by
  \[
    C^{(j)} := \set*{x\in \mcal X: g^{(j)} = \argmin_{g\in G} \kappa(\pi(\tilde x), g^{(j)})^p},
  \]
  we have
  \[
    \cost(\mcal C)\leq (1+\varepsilon) \beta \OPT.
  \]
\end{thm}

\begin{pf}
  We have
  \begin{align*}
    \cost(\mcal C)
    &\leq (1+\varepsilon) \widetilde{\cost}(\mcal C) &&\text{\Cref{prop:discretization cost preservation}} \\
    &\leq (1+\varepsilon)^2 \widetilde{\cost}(\pi(\mcal C)) &&\text{\Cref{prop:weighted dim reduction}}\\
    &\leq (1+\varepsilon)^2 \widetilde{\cost}(G) &&\text{$G$ induces $\pi(\mcal C)$} \\
    &\leq (1+\varepsilon)^2 \beta \widetilde{\OPT}_\pi &&\text{$G$ is $\beta$-approximate} \\
    &\leq (1+\varepsilon)^3 \beta \OPT. &&\text{\Cref{cor:discretization dim reduction opt preservation}}
  \end{align*}

  Choosing $\varepsilon'$ to be a constant fraction of $\varepsilon$ concludes the proof.
\end{pf}

\subsection{Summary of Dimensionality Reduction}\label{apx:outline dim reduction}
Before proving \Cref{thm:euclidean clustering algorithm},
let us summarize the stages of our pipeline.
\begin{enumerate}[1)]
    \item Scale the input data by $\nicefrac1{\sqrt d}$.
    \item Estimate $\OPT$ of the scaled data within a multiplicative error.
    \item Discretize the scaled data with a fine grid.
    \item Project the scaled data with a random map $\pi\in \pi_{d, m}$.
    \item Compute a coreset on the projected data.
    \item Estimate the probability mass at each point of the coreset.
    \item Call the $\beta$-approximation oracle to produce $k$ centers $g^{(1)}, \dots, g^{(k)}$.
    \item Output the clustering function induced by the centers.
\end{enumerate}

The clustering function proceeds as follows.
Given a point $x\in\mcal X$,
\begin{enumerate}[1)]
    \item Scale $x$ by $\nicefrac1{\sqrt d}$.
    \item Discretize the scaled point with the same grid used in the algorithm.
    \item Project the discretized point using the same $\pi$ used in the algorithm.
    \item Output the label of the closest center $g^{(j)}$ in the lower dimension for $j\in [k]$.
\end{enumerate}

We are now ready to prove \Cref{thm:euclidean clustering algorithm},
which we restate below for convenience.

\kpClusteringEuclidean*

\begin{pf}[\Cref{thm:euclidean clustering algorithm}]
    We randomly draw some $\pi\in \pi_{d, m}$ for
    \[
        d = O\left( \frac{p^4}{\varepsilon^2} \log \frac{k}{\varepsilon\delta} \right).
    \]
    Then,
    we discretize a sufficiently large sample after scaling
    and push it through $\pi$.
    We then run the corresponding algorithm (cf. \Cref{thm:statistical k-medians algorithm}, \Cref{thm:statistical k-means algorithm}) to produce some centers $G$ for the compressed data.
    This allows us to partition $\mcal X$ according to the closest member of $G$ in the lower dimension.
    An application of \Cref{thm:dim reduction} yields the performance guarantees of this algorithm.

    The sample complexity guarantee from \Cref{thm:statistical k-medians algorithm}, \Cref{thm:statistical k-means algorithm} is for the scaled distribution
    and the scaled optimal cost satisfies $\OPT_\pi' = d^{-\nicefrac p2} \OPT_\pi$.
    Hence the sample complexity resolves to
    \begin{align*}
      \tilde O \left( \poly\left( \frac{kd}{\rho \OPT} \right) \left( \frac{2\sqrt m}{\varepsilon} \right)^{O(m)}\log\frac1\delta \right)
    \end{align*}
    
    We conclude the proof by remarking that for the Euclidean metric,
    $\Delta = \sqrt d$
    and that $O(\log n)^{O(\log n)} = n^{O(\log \log n)}$.
\end{pf}

\section{\texorpdfstring{$k$}{k}-Centers}\label{sec:overview k-centers}
In this section,
we will explain our approach to solving the statistical
$k$-centers problem (cf. \Cref{prob:stat k-centers}). As we explained before, since this problem has a different flavor compared
to $(k,p)$-clustering, we introduce some extra assumptions
in order to be able to solve it from samples. To the
best of our knowledge, this is the first work
that considers a model of statistical flavor for the
$k$-centers problem.
\subsection{Assumptions}
First, we describe an assumption that is necessary to 
obtain good clustering solutions even if we do not insist on
using replicable algorithms.

\begin{assum}[Clusterable]\label{assum:clusterable}
For some absolute constants $\beta, B > 0$,
an i.i.d. sample $S = x_1,\ldots,x_n$ of size $n$ from $\P$ has the following property.
With probability at least $q$,
there exists some $(\beta, B)$-approximate solution $F_*$
to \Cref{prob:stat k-centers}
such that, for every $f\in F_*$ there is some $x_f \in S$ with $F_*(x_f) = f$.
Here $q\in [0, 1]$.
\end{assum}
\Cref{assum:clusterable} is necessary to obtain good clustering solutions even in the non-replicable case.
Our distribution $\P$ must be sufficiently well-behaved
so that solving the sample $k$-centers problem translates to a good solution in the population case.
Essentially,
this assumption states that with high probability,
the sample will contain a point from every
single cluster of \emph{some} $(\beta, B)$-approximate solution. 
In order to design \emph{replicable} algorithms, we need to make a stronger assumption.

\begin{assum}[Replicable]\label{assum:reproducible}
  For some absolute constants $\beta, B$,
  there exists a $(\beta, B)$-approximate solution $F_*$ to \Cref{prob:stat k-centers}
  with the following property.
  With probability at least $q$, in an i.i.d. sample $S = x_1,\ldots,x_n$
  of size $n$ from $\P$, for every $f\in F_*$,
  there is some $x_f \in S$ with $F_*(x_f) = f$.
  Here $q\in [0, 1]$.
\end{assum}

Note that \Cref{assum:clusterable} guarantees that, with high probability,
every sample will help us identify \emph{some} good solution.
However, these good solutions could be vastly different when we observe
different samples, thus this assumption is not sufficient to design a 
replicable algorithm with high utility.
\Cref{assum:reproducible} is necessary to attain replicable solutions.
Indeed, this assumption essentially states that there is a \emph{fixed} solution
for which we will observe samples from every cluster with high probability. If this does not hold and we observe points from 
clusters of solutions that vary significantly, we cannot hope to replicably
recover a fixed approximate solution.
Note that \Cref{assum:reproducible} is stronger than \Cref{assum:clusterable}, since we have flipped the order of the 
existential quantifiers.
One way to satisfy this assumption is to require that 
there is some $(\beta, B)$-approximate solution 
so that the total
mass that is assigned to every cluster of it
is at least $\nicefrac1n$ for some $n \in \mathbb{N}.$ Then,
by observing $\widetilde{O}(n)$ points,
we will receive at least
one from every cluster with high probability. This is made formal
in \Cref{prop:sampling}.

\begin{prop}[\citep{janson2018tail}]\label{prop:sampling}
  Let $\delta \in (0,1)$. Under \Cref{assum:reproducible},
  a sample of
  \[
    O\left( \frac{nm}{q} \log\frac1\delta \right)
  \]
  points contains
  at least $m$ points from each cluster of $F_*$
  with probability at least $1- \delta$.
\end{prop}

\begin{pf}[\Cref{prop:sampling}]
  Let $Y_i$ be a geometric variable with success rate $q$,
  which denotes the number of batches of $n$ samples we draw until a point from each cluster of $F_*$ is observed.
  Then $Y := \sum_{i=1}^m Y_i$ upper bounds the number of batches until we obtain $m$ points from each cluster.

  As shown in \citet{janson2018tail},
  \begin{align*}
    \P\set{Y\geq \lambda \E Y} &\leq \exp(1-\lambda) \\
    \E Y &= \frac{m}{q}. 
  \end{align*}
  The result follows by picking $\lambda = 1 + \log(\nicefrac1\delta).$
\end{pf}

\subsection{High-Level Overview of the Approach}
We are now ready to provide a high-level overview of our approach
to derive \Cref{thm:informal k-centers sample complexity of reproducible active cells}. We first take a fixed grid of side $c$
in order to cover the unit ball. Then, we sample sufficiently many points
from $\P$. Subsequently, we round all the points of the sample 
to the centers of the cells of the grid that they fall into. Using these points,
we empirically estimate the density of the distribution on every cell of the grid.
Next, in order to ensure that our solution is replicable, we take a random 
threshold from a predefined interval and discard the points from
all the cells whose density falls below the threshold. Finally, we call
the approximation oracle that we have access to using the points that have
survived after the previous step. 
We remark that unlike the $(k,p)$-clustering problem
(cf. \Cref{prob:stat cluster}), to the best of our knowledge,
there does not exist
any dimensionality reduction techniques that apply to the $k$-centers
problem. In
the following subsections we explain
our approach in more detail.

\subsection{Oracle on Sample}
As we explained before, we require black-box access to
an oracle $\O$ for the combinatorial $k$-centers problem on a sample
which outputs $(\wh \beta, \wh B)$-approximate solutions $\wh F$ for \Cref{prob:k-centers}. Thus, we need to show that, with high 
probability, the output of this solution is a good approximation
to \Cref{prob:stat k-centers}. This is shown in \Cref{prop:k-centers-output of oracle without grid}.

\begin{prop}\label{prop:k-centers-output of oracle without grid}
  Suppose there is a $(\beta, B)$-approximate solution $F_*$ for \Cref{prob:stat k-centers}
  and that we are provided with a sample of size $ O (n \log(\nicefrac1\delta)/q)$
  containing an observation from each cluster of $F_*$, with probability at least $1-\delta$.
  Then, the output $\wh F$ of the oracle $\O$ on the sample satisfies
  \[
    \kappa(x, \wh F) \leq (2\beta + \wh \beta) \OPT + 2B + \wh B.
  \]
  for all $x\in \mcal X,$ with probability
  at least $1-\delta$.
\end{prop}

In the previous result $(\beta, B),(\wh \beta, \wh B)$ 
are the approximation parameters that
we inherit due to the fact that we receive samples from $\P$ and that
we solve the combinatorial
problem approximately using $\O$, respectively.

\begin{pf}[\Cref{prop:k-centers-output of oracle without grid}]
      Let $F_*$ denote the $(\beta, B)$-approximate solution from \Cref{assum:reproducible}. We condition on the event that such
  a solution exists in the sample, which occurs with  probability
  at least $1-\delta$.

  Fix $x\in \mcal X$, which is not necessarily 
  observed in the sample.
  We pay a cost of $\beta\OPT + B$ to travel to the nearest center of $F_*$.
  Next, we pay another $\beta \OPT + B$ to travel to the observed sample from this center.
  Finally, we pay a cost of
  \begin{align*}
    \wh\cost(\wh F)
    &\leq \wh \beta \wh\OPT + \wh B \\
    &\leq \wh \beta \OPT + \wh B.
  \end{align*}
  to arrive at the closest center in $\wh F$.
\end{pf}

\subsection{Oracle with Grid}
The issue with using directly the output of $\O$ on a given sample is 
that the result will not be replicable, since we have no control over 
its behavior. In a similar way as with the $(k,p)$-clustering problem, 
the main tool that we have in our disposal to ``stabilize'' the output of $\O$
is to use a grid. Unlike the other problems, we cannot use the hierarchical grid
approach in this one. The reason is that, due to the min-max nature of k-centers,
we need to cover the whole space and not just the points where the distribution
is dense. Recall that so far we have established that the
oracle $\O$ outputs $(2\beta + \wh \beta, 2B + 2 \wh B)$-approximate solutions $F$ for \Cref{prob:stat k-centers},
given that \Cref{assum:reproducible} holds.
\Cref{alg:oracle grid} ``discretizes'' the points of the sample
using a grid with side-length $c$ before applying the oracle.
\Cref{prop:k-centers oracle on grid} shows that by doing that, 
we have to pay an extra additive term that is of the order $O(c \Delta).$

\begin{algorithm}[h]
\caption{Oracle with Grid}\label{alg:oracle grid}
\begin{algorithmic}[1]
  \STATE {\bfseries Oracle with Grid}(oracle $\O$, sample $x_1, \dots, x_n$, grid length $c$):
  \FOR{$i \gets 1, \dots, n$}
    \STATE Decompose $x_i = z_i\cdot c + x_i'$ for some $z_i\in \Z^d, x_i'\in [0, c)^d$
    \STATE $\tilde x_i \gets z_i\cdot c + \left( \nicefrac{c}2, \nicefrac{c}2, \dots, \nicefrac{c}2 \right)$ \COMMENT{$\tilde x_i$ is the center of the cell that contains $x_i.$}
  \ENDFOR
  \STATE Return $\O(\tilde x_1, \dots, \tilde x_n)$
\end{algorithmic}
\end{algorithm}

\begin{prop}\label{prop:k-centers oracle on grid}
  Suppose there is a $(\beta, B)$-approximate solution $F_*$ for \Cref{prob:stat k-centers} and that we are provided with a sample of size $n$
  containing an observation from each cluster of $F_*$. 
  Let $\tilde G$ denote the output of \Cref{alg:oracle grid} using a $(\hat{\beta},\hat{B})-$approximation
  oracle $\mcal O$ for \Cref{prob:k-centers}.
  Then, for any $x\in \mcal X$, 
  \[
    \kappa(x, \tilde G)
    \leq (2\beta + \wh \beta) \OPT + 2B + \wh B + (4\beta + 2\hat \beta + 1)c\Delta \,.
  \]
\end{prop}

\begin{pf}[\Cref{prop:k-centers oracle on grid}]
      For the sake of analysis,
  imagine we used the grid to discretize all points of $\mcal X \mapsto \tilde{\mcal X}$.
  Then $F_*$ is a solution with cost $\beta \widetilde\OPT + B + c\Delta$
  and we observe a point from each of its clusters in our discretized sample.
  Here $\widetilde\OPT$ is the cost of a solution to \Cref{prob:stat k-centers} on $\tilde{\mcal X}$.

  \Cref{prop:k-centers-output of oracle without grid} shows that the oracle returns a solution $\widetilde G$ with cost at most
  \[
    (2\beta + \wh \beta) \widetilde\OPT + (2B + c\Delta) + \wh B
  \]
  on the discretized points.

  Then for any $x\in \mcal X$,
  \begin{align*}
    \widetilde \OPT
    &= \kappa(x, \tilde G_{\OPT}) \\
    &\leq \kappa(x, \tilde x) + \kappa(\tilde x, \tilde G_{\OPT}) \\
    &\leq \kappa(x, \tilde x) + \kappa\left( \tilde x, F_{\OPT} \right) \\
    &\leq 2\kappa(x, \tilde x) + \kappa(x, F_{\OPT}) \\
    &\leq 2c\Delta + \OPT.
  \end{align*}

  The conclusion follows by combining the two inequalities.
\end{pf}

\subsection{Replicable Active Cells}
At a first glance,
we might be tempted to work towards replicability by arguing that, with
high probability, every grid cell is non-empty in one execution of the algorithm
if and only if it is non-empty in another execution.
However, since we have no control over $\P$, it is not clear
how one can prove such a statement.
Our approach is to take more samples and ignore cells which do not have a sufficient number of points using some random threshold. 
The idea is similar to the heavy-hitters algorithm (cf. \Cref{alg:reproducible heavy hitters})
originally conceived by \citet{impagliazzo2022reproducibility}
and is presented in \Cref{alg:reproducible active cells no grid}.
Then,
we move all the points of active cells to the center of the cell.
Notice that in the $k$-center objective, unlike the $(k,p)$-clustering
objective,
we do not care about how much mass is placed in a cell,
just whether it is positive or not.

We first derive a bound on the number of cells that the distribution puts mass on. It is not hard to see that since we are working in 
a bounded domain,
the grid contains at most $(\nicefrac1c)^d$ cells in total.
Thus there are at most $M \leq (\nicefrac1c)^d$ cells that have non-zero mass.
\Cref{prop:k-centers number of cells that intersect with cluster}
shows how we can obtain such a bound in the unbounded domain setting.

\begin{prop}\label{prop:k-centers number of cells that intersect with cluster}
 Let $F_*$ be a $(\beta, B)$-approximate solution for \Cref{prob:stat k-centers}. Let $M$ denote the maximum number of cells which intersect some cluster of $F_*$ and
$\tilde \P$ the discretized distribution that is supported on centers of the cells of
the grid of size $c$. Then, the support of $\tilde \P$ is at most $kM$ where
  \[
    M \leq \left( \frac{c + 2c\Delta + 2 (\beta \OPT + B)}{c} \right)^d.
  \]
\end{prop}

\begin{pf}[\Cref{prop:k-centers number of cells that intersect with cluster}]
    Fix a center $f$
  and consider the cell $Z$ containing the center.
  The farthest point in the farthest cell in the cluster of $f$
  is of distance at most $\beta \OPT + B + c\Delta$
  and thus the cells we are concerned with sit in an $\ell_\infty$ ball of that radius.
  Not including $Z$,
  we walk past at most
  \[
    \frac{\beta \OPT + B + c\Delta}{c}
  \]
  cells along the canonical basis of $\R^d$,
  hence there are at most
  \[
    \left[ 1 + 2c\Delta + \frac{2}{c}(\beta \OPT + B) \right]^d
  \]
  such cells.
\end{pf}

Notice that the previous discussion and \Cref{prop:k-centers oracle on grid} 
reveal a natural
trade-off in the choice of the parameter $c$: by reducing $c$,
we decrease the additive error of our approximation algorithm but 
we increase the number of samples at a rate $(\nicefrac1c)^d.$
Since we have bounded the number of cells from which we can observe a point,
the next step is to use a sufficiently large number of samples so that
we can estimate replicably whether a cell is active or not.
This is demonstrated in \Cref{alg:reproducible active cells no grid}.
\begin{algorithm}[h]
\caption{Replicable Active Cells}\label{alg:reproducible active cells no grid}
\begin{algorithmic}[1]
  \STATE {\bfseries rActiveCells}(grid length $c$):
  \STATE $m \gets \tilde O\left( \nicefrac{\lambda knM^2}{\rho^2} \right)$
  \STATE $N \gets \lambda nmM$
  \STATE Sample $x_1, \dots, x_N \sim \P$, 
  \STATE Lazy initialize counter $Z = 0$ for every cell of the grid with length $c$
  \FOR{$i \gets 1, \dots, N$}
    \STATE $G_i \gets $ cell that samples $x_i$ falls into
    \STATE Increment $Z(G_i) \gets Z(G_i) + 1$
  \ENDFOR
  \STATE Choose $v\in \left[ 0, \nicefrac{m}{N} \right]$ uniformly randomly
  \STATE Output all $G_i$ such that $\nicefrac{Z(G_i)}{N} \geq v$
\end{algorithmic}
\end{algorithm}

\begin{prop}\label{prop:k-centers sample complexity of just reproducible
active cells}
      Suppose \Cref{assum:reproducible} holds and let $F_*$ be the $(\beta,B)$-approximate solution. Let $\delta, \rho \in (0,1)$.
  Then
  \Cref{alg:reproducible active cells no grid} is $\rho$-replicable
  and returns cells of the grid
  such that at least one sample from each cluster of $F_*$
  falls into these cells with probability at least $1-\delta.$
  Moreover,
  it has sample complexity
  \[
    \tilde O\left( \frac{n^2kM^3\log(\nicefrac1\delta)}{q^2\rho^2}\ \right).
  \]
\end{prop}

\begin{pf}[\Cref{prop:k-centers sample complexity of just reproducible
active cells}]
  Enumerate the $kM$ cells within the support of $\P$
  and let the counter $Z^{(j)}$ denote the number of samples at the $j$-th cell.
  Then $Z = (Z^{(1)}, \dots, Z^{(kM)})$ follows a multinomial distribution
  with parameters $(p^{(1)}, \dots, p^{(kM)})$ and $N,$ where $p$ is unknown to us.
  Nevertheless, by \Cref{prop:BHC inequality},
  for any $\varepsilon\in (0, 1)$,
  sampling
  \[
    N \geq \frac{\ln\nicefrac5\delta + kM\ln 2}{2\varepsilon^2}
  \]
  points implies that $\sum_{j=1}^a \abs{\wh p^{(j)} - p^{(j)}} < 2\varepsilon$
  with probability at least $1-\nicefrac\delta5$.
  From the definition of $N$,
  this is equivalent to
  \[
    m\geq \frac{\ln\nicefrac5\delta + kM\ln 2}{ nM\varepsilon^2}.
  \]

  By \Cref{prop:sampling},
  we observe at least $mM$ points from each cluster of $F_*$
  with probability at least $1-\delta$.

  Let $\wh p_1^{(j)}, \wh p_2^{(j)}$ be the empirical mean of $Z^{(j)}$
  across two different executions of \Cref{alg:reproducible active cells no grid}.
  Then by the triangle inequality,
  $\sum_{j=1}^{kM} \abs{\wh p_1^{(j)} - \wh p_2^{(j)}} \leq 4\varepsilon$.
  
  Since we observe at least $Mm$ points from each cluster of $F_*$,
  there is at least one point from each cluster
  with at least $m$ observations.
  Thus outputting the points $j$ such that $\wh p^{(j)} \geq v$ for some $v\in \left[ 0, \nicefrac{m}{N} \right]$
  means we always output one point from each cluster. 

  Now, the outputs between two executions are not identical
  only if the point $v$ ``splits'' some pair $\wh p_1^{(j)}, \wh p_2^{(j)}$.
  This occurs with probability at most $\nicefrac{4\varepsilon}{(m/N)} = \nicefrac{4N\varepsilon}{m}$,
  To bound this value by $\nicefrac\rho5$,
  set
  \begin{align*}
    \varepsilon
    &\leq \frac{\rho m}{20 N} \\
    &= \frac{\rho}{20\lambda nM}.
  \end{align*}
  Plugging this back yields
  \[
    m
    \geq \frac{400\lambda nM\ln\frac5\delta + 400\lambda nkM^2 \ln 2}{\rho^2}.
  \]

  All in all,
  \Cref{alg:reproducible active cells no grid} outputs the desired result
  with probability at least $1-\delta$ and is $\rho$-replicable
  given
  \[
   \tilde O\left( \frac{n^2kM^3\log(\nicefrac1\delta)}{q^2\rho^2}\ \right) 
  \]
  samples from $\P$.
\end{pf}

\subsection{Putting it Together}
We now have all the ingredients in place to prove \Cref{thm:informal k-centers sample complexity of reproducible active cells}. The proof follows directly by 
combining \Cref{prop:k-centers oracle on grid}, and \Cref{prop:k-centers sample complexity of just reproducible
active cells}. Essentially, we replicably estimate the active cells
of our grid and then move all the points of each active cell to its
center.

For completeness, we also state the formal version of \Cref{thm:informal k-centers sample complexity of reproducible active cells}.
\begin{thm}[\Cref{thm:informal k-centers sample complexity of reproducible active cells}; Formal]
\label{thm:formal k-centers sample complexity of reproducible active cells}
  Let $\delta, \rho \in (0,1)$. Suppose \Cref{assum:reproducible} holds and let $F_*$ be the $(\beta,B)$-approximate solution. 
  Then,
  given black-box access to a $(\hat{\beta}, \hat{B})$-approximation oracle for \Cref{prob:k-centers},
  \Cref{alg:reproducible active cells no grid} is $\rho$-replicable
  and returns a solution $\hat{F}$ such that
  \[
    \kappa(x, \hat{F})
    \leq (2\beta + \wh \beta) \OPT + 2B + \wh B + (4\beta + 2\hat \beta + 1)c\Delta
  \]
  with probability at least $1-\delta$.
  Moreover,
  the algorithm requires at most
  \[
    \tilde O\left( \frac{ n^2k\left(\nicefrac1c\right)^{3d}\log(\nicefrac1\delta)}{q^2\rho^2} \right)
  \]
  samples from $\P$.
\end{thm}

\end{document}